%% file: uai2025-template.tex
\documentclass[accepted]{uai2025} 
                        
\usepackage{natbib} 
\usepackage{mathtools} 
\usepackage{booktabs} 

\usepackage{forloop}
\usepackage{url}

\usepackage{amsmath}
\usepackage{amssymb}
\usepackage{mathtools}
\usepackage{amsthm}
\usepackage{authblk}
\usepackage{bbding}
\usepackage{multirow}
\usepackage{comment}
\usepackage[table,dvipsnames]{xcolor}
\usepackage{hyperref}
\hypersetup{
    colorlinks=true,
    linkcolor=red,
    filecolor=magenta,      
    urlcolor=cyan,
    citecolor=Blue,
    pdftitle={Overleaf Example},
    pdfpagemode=FullScreen,
    }
\theoremstyle{plain}
\newtheorem{theorem}{Theorem}[section]

\theoremstyle{definition}

\newtheorem{assumption}[theorem]{Assumption}
\theoremstyle{remark}

\newcommand*\dif{\mathop{}\!\mathrm{d}}

\title{\texttt{LucidAtlas}: Learning Uncertainty-Aware, Covariate-Disentangled, Individualized Atlas Representations}

\author[1]{Yining Jiao}
\author[1]{Sreekalyani Bhamidi}
\author[1]{Huaizhi Qu}
\author[1]{Carlton Zdanski}
\author[1]{Julia Kimbell}
\author[1]{Andrew Prince}
\author[1]{Cameron Worden}
\author[1]{Samuel Kirse}
\author[1]{Christopher Rutter}
\author[1]{Benjamin Shields}
\author[1]{Jisan Mahmud}
\author[1]{Tianlong Chen}
\author[2]{Marc Niethammer}
\affil[1]{UNC-Chapel Hill} \affil[2]{UCSD}

\begin{document}
\maketitle

\begin{abstract}
 The goal of this work is to develop principled techniques to extract information from high dimensional data sets with complex dependencies in areas such as medicine that can provide insight into individual as well as population level variation. We develop \texttt{LucidAtlas}, an approach that can represent spatially varying information, and can capture the influence of covariates as well as population uncertainty. As a versatile atlas representation, \texttt{LucidAtlas} offers robust capabilities for covariate interpretation, individualized prediction, population trend analysis, and uncertainty estimation, with the flexibility to incorporate prior knowledge. Additionally, we discuss the trustworthiness and potential risks of neural additive models for analyzing dependent covariates and then introduce a marginalization approach to explain the dependence of an individual predictor on the models' response (the atlas). To validate our method, we demonstrate its generalizability on two medical datasets. Our findings underscore the critical role of by-construction interpretable models in advancing scientific discovery. Our code will be publicly available upon acceptance. 
\end{abstract}

\section{Introduction}
\label{sec.intro}


In the context of medical image analysis and computational anatomy, an \emph{atlas} is a standardized representation of biological structures that serves as a reference model~\citep{joshi2004unbiased, thompson2002framework}. An atlas is often created by aggregating data from multiple patients to represent "typical" or "average" anatomy. Atlases are crucial in medical research, diagnosis, and treatment planning, providing a baseline for comparison with individual patient data~\citep{atlashong2013pediatric,commowick2005incorporating}. Although often representing average structures, some atlases also incorporate information on anatomical variability within a population~\citep{jin2019howmany,kovavcevic2005three}.

The goal of this work is to enhance atlas representations by incorporating covariates and uncertainties, providing clinicians with a more comprehensive tool for disease analysis. Our approach goes beyond traditional models, offering covariate disentanglement and uncertainty estimation for improved population understanding. Specifically, accounting for covariates and capturing subject- and population-level aspects about the data is important to address the following questions relevant for clinicians and technical users who might want to build upon our model: 



\paragraph{Covariate-Level} 
    \textcircled{1}  \emph{Covariate Disentanglement.}  Understanding the effects of covariates is frequently a goal of medical studies. Therefore, it is critical to disentangle covariate effects on the population to interpret an atlas representation and to ensure that these effects align with human knowledge. Inherently interpretable models are desirable for atlas building because such atlas representations are then transparent by construction~\citep{rudin2019stop}.
    \textcircled{2} \emph{Covariate Marginalization}. Clinicians usually understand the importance of a covariate by analyzing its relationship with the response, e.g.,  how a population trend evolves under the control of a specic covariate regardless of the existence of other covariates. For example, how does a brain change for normal people versus Alzheimer patients, regardless of age?
    \textcircled{3}  \emph{Prior Knowledge.} Suppose we have  prior knowledge about some or all covariates, e.g., monotonicity\footnote{For example, a pediatric airway should typically expand with age, as a child grows.} how can such prior knowledge assist in better atlas construction? Do our modeling assumptions  align with our prior knowledge and are they reasonable for human data?
\paragraph{Subject-Level}
    \textcircled{1}  \emph{Predictions based on the Population Trend.} Is the constructed atlas a good predictor for individual anatomies given their covariates?  \textcircled{2} \emph{Individualized Temporal Analysis.} Can the model provide individualized predictions at time $B$ based on observations based on a prior time $A$? This is a challenging task, as the collected data often consists of numerous one-time observations rather than longitudinal data.
\paragraph{Population-Level} 
    \textcircled{1} \emph{Individual Variability.}  Variation across the population may surpass the variation explained by covariates. How can we quantify population variation not captured by the covariates? \textcircled{2} \emph{Heteroscedasticity.}  Variability may be heteroskedastic across covariates and anatomical geometry. For example, population variance may differ change with age and airway location when constructing a pediatric airway atlas. 
\paragraph{Spatial-Level} 
    \textcircled{1} \emph{Spatial Dependence.} Capturing spatial dependence is essential for atlas construction. For instance, different anatomical locations may exhibit distinct population trends and variations influenced by covariates, highlighting the need to model spatial dependence effectively.
    

To address covariate-, subject-, population-, and spatial-questions simultaneously, we propose the versatile \texttt{LucidAtlas} model. This uncertainty-aware, covariate-disentangled individualized atlas representation extends the Neural Additive Model (NAM)~\citep{agarwal2020neural} to enhance atlas construction by integrating uncertainty quantification and incorporating prior knowledge. We introduce a marginalization approach that incorporates covariate dependencies to provide a more comprehensive covariate interpretation; while NAM's interpretations only focus on individual covariate effects in isolation. 

The main contributions of \texttt{LucidAtlas} are as follows:
\begin{itemize}
    \item[1)] \texttt{LucidAtlas} is a versatile atlas representation capable of enhancing traditional atlas representations by incorporating covariates, uncertainties, and prior knowledge, providing clinicians with a more comprehensive tool for disease analysis.
    \item[2)] We show potential risks when using NAMs for covariate interpretation in the context of dependent features. We further propose a marginalization approach to address these shortcomings. 
    \item[3)] We validate our approach on two medical datasets
    : 1) the OASIS Brain Volume dataset~\citep{jack2008alzheimer}, and 2) a Pediatric Airway Shape dataset. Our experiments quantitatively demonstrate the superior performance of \texttt{LucidAtlas} compared to baseline methods.
\end{itemize}


\begin{table}[t]
\centering
\resizebox{0.49\textwidth}{!}{%
\begin{tabular}{l|cc|c|c|c}
\hline
 &
  \multicolumn{2}{c|}{\cellcolor[HTML]{F8E5E4}Covariate-} &
  \cellcolor[HTML]{E2E3F8}Subject- &
  \cellcolor[HTML]{FFFFC7}Population- &
  \cellcolor[HTML]{E9FEE9}Spatial- \\ \cline{2-6} 
\multirow{-2}{*}{Method} &
  \multicolumn{1}{c|}{Cov. Marg.} &
  Prior. &
  Ind. Pred. &
  Hetero.+Aleatoric &
  Spa. Dep. \\ \hline
NAM~\citep{agarwal2020neural}      & \multicolumn{1}{c|}{\XSolidBrush} & \XSolidBrush & \XSolidBrush & \XSolidBrush & \XSolidBrush \\ \hline
OAK~\citep{lu2022oak}              & \multicolumn{1}{c|}{\Checkmark}   & \XSolidBrush & \XSolidBrush & \XSolidBrush & \XSolidBrush \\ \hline
LA-NAM~\citep{bouchiat2023lanam}   & \multicolumn{1}{c|}{\XSolidBrush} & \XSolidBrush & \Checkmark   & \XSolidBrush & \XSolidBrush \\ \hline
NAMLSS~\citep{thielmann2024namlss} & \multicolumn{1}{c|}{\XSolidBrush} & \XSolidBrush & \Checkmark   & \Checkmark   & \XSolidBrush \\ \hline
NAISR~\citep{jiao2023naisr}        & \multicolumn{1}{c|}{\XSolidBrush} & \XSolidBrush & \Checkmark   & \XSolidBrush & \Checkmark   \\ \hline
\texttt{LucidAtlas}(Ours)          & \multicolumn{1}{c|}{\Checkmark}   & \Checkmark   & \Checkmark   & \Checkmark   & \Checkmark   \\ \hline
\end{tabular}
}
\caption{\small Comparison of interpretable representations based on the desirable properties discussed in Sec.\ref{sec.intro}. \textbf{Cov.Marg.} denotes covariate marginalization. \textbf{Prior.} indicates prior knowledge. \textbf{Ind. Pred.} indicates individualized prediction, i.e., whether the model can predict a response for time $B$ given an earlier observation at time $A$. \textbf{Hetero.+Aleatoric} indicates whether the model considers heteroscedasticity when modeling aleatoric uncertainty. \textbf{Spa. Dep.} indicates spatial dependence. A \Checkmark indicates that a model has a property; a \XSolidBrush indicates that it does not. Only \texttt{LucidAtlas} has all the desired properties. }
\label{tab.lit}
\end{table}

\section{Related Work}
We first introduce the three most related research directions. 

\paragraph{Additive Models}
Model-agnostic methods, such as Partial Dependence~\citep{friedman2001greedy}, SHAP~\citep{lundberg2017unified}, and LIME~\citep{ribeiro2016should}, offer a standardized approach to explaining machine learning predictions. However, when applied to deep neural networks (DNNs), these methods may fail to provide faithful representations of their full complexity~\citep{rudin2019stop}. A more transparent alternative involves leveraging Generalized Additive Models (GAMs)~\citep{hastie2017generalized}, where the response variable \( y \) is modeled using an additive structure:  
\begin{small}
\begin{equation}
E[y | c_1, . . . , c_N] = h(\beta_0 + f_1(c_1) + \dots + f_N(c_N))
\label{eq.nam}
\end{equation}
\end{small}
where \( h(\cdot) \) is the inverse of the link function (a form of activation function) ; $\beta_0$ denotes the intercept and $f_{i}(\cdot)$ represent independent functions for the $i^{th}$ covariate. Neural Additive Models (NAMs)~\citep{agarwal2020neural, jiao2023naisr} build upon this framework, offering enhanced interpretability while maintaining the flexibility of neural networks. Specifically, for NAMs the functions $f_i(\cdot)$ are deep neural networks. NAISR~\citep{jiao2023naisr} pioneers the use of NAMs to capture spatial deformations with respect to an estimated atlas shape that is modulated by covariates. \emph{\texttt{LucidAtlas} extends this concept by integrating NAMs to construct an atlas that captures population trends and uncertainties with spatial dependencies}.

\paragraph{Epistemic Uncertainty versus Aleatoric Uncertainty}

Estimating uncertainty is important to understand the quality of a model fit and to capture variations across the data population. Two different types of uncertainties need to be distinguished: epistemic uncertainty captures model uncertainty whereas aleatoric uncertainty captures uncertainty in the data~\citep{hullermeier2021survey}. 

More attention is generally paid to epistemic uncertainties in the context of interpretable models~\citep{wang2025uncertaintysurvey}. NAMs used ensembling to estimate model uncertainties~\citep{agarwal2020neural}. LA-NAM used a Laplace approximation for uncertainty estimation~\citep{bouchiat2023lanam} with NAMs. In atlas construction, aleatoric uncertainty is especially important when individual differences in a dataset are large. Capturing aleatoric uncertainty is crucial in medicine to understand population variations. NAMLSS~\citep{thielmann2024namlss} can model aleatoric uncertainty by using NAMs to approximate the parameters $\{\theta^{(n)}\}$ of a chosen data distribution~\citep{thielmann2024namlss}, as
\begin{small}
\begin{equation}
\theta^{(n)}=h^{(n)}\left(\beta^{(n)}+\sum_{i=1}^{N} f_{i}^{(n)}\left(c_{i}\right)\right) 
\label{eq.namlss}
\end{equation}
\end{small}
where $\theta^{(n)}$ can for example be the mean and variance of Gaussian distributions; $\beta^{(n)}$ denotes the parameter-specific intercept and $f_{i}^{(n)}$ represents the feature network for the $n$-th parameter for the $i$-th feature. \emph{\texttt{LucidAtlas} extends NAMLSS to a more versatile representation, enabling individualized prediction, incorporating prior knowledge, and capturing spatial dependencies.}

\paragraph{Heteroscedasticity versus Homoscedasticity}
Distinguishing between homoscedasticity and heteroscedasticity is crucial in statistical analysis, especially for regression models. Homoscedasticity indicates constant variance of random variables , whereas heteroscedasticity indicates that the variance of random variables may differ~\citep{wooldridge2016homohetr}. For example, when modeling airway cross-sectional area the population variance may change (increase) with age. \texttt{LucidAtlas} assumes and supports estimating heteroscedasticity with respect to different locations in an anatomical region and with respect to covariates across a patient population. Many interpretable approaches assume homoscedasticity, e.g., OAK-GP~\citep{lu2022oak} and LA-NAM~\citep{bouchiat2023lanam} assume homoscedasticity in their additive networks. To our knowledge, only NAMLSS considers heteroscedasticity in its additive network design~\citep{thielmann2024namlss}. However, NAMLSS interprets individual covariate effects and uncertainties in isolation resulting, as we will see, in difficulties for data interpretation. \emph{\texttt{LucidAtlas} advances beyond NAMLSS by capturing spatial heteroscedasticity and incorporating covariate dependencies via a marginalized covariate interpretation approach.}

Table~\ref{tab.lit} compares \texttt{LucidAtlas} to related interpretable models with respect to the discussed properties above. A more comprehensive discussion of
related work is available in Sec.~\ref{supp.sec.supp_related_work} of the Supplementary Material.

\section{Method}
\label{sec.method_model_arch}

Sec.~\ref{sec.prob_des} introduces our general atlas construction formulation. Sec.~\ref{sec.workflow_lucidatlas} describes \texttt{LucidAtlas}, which by construction ensures \emph{covariate disentanglement based on an additive model formulation}. Sec.~\ref{subsec.why_mrg} discusses the potential concerns for neural additive models when dealing with dependent covariates. To address these concerns  Sec.~\ref{subsubsec.how_marg} introduces our covariate marginalization approach. Table~\ref{supp.tab.exp_notations} lists the mathematical notations used in this paper. 

\subsection{General Atlas Formulation}
\label{sec.prob_des}
Consider a set of anatomies $\{\mathcal{Y}^k\}$, where each anatomy $\mathcal{Y}^k$ is associated with a vector of covariates $\boldsymbol{c}^k = [c^k_1, \dots, c^k_i, \dots, c^k_N]$. A point within an anatomy is denoted as $x$, with an observed value $y \in \mathbb{R}$, which is the primary focus of our study.

Our goal is to construct an atlas representation that addresses the questions outlined in Section~\ref{sec.intro} by learning the mapping from covariates $\boldsymbol{c}$ and spatial location $x$ to observations $y$, while simultaneously accounting for heteroscedastic uncertainties within the population, represented as $p(y \mid \boldsymbol{c}, x) = \mathcal{N}(f^m(\boldsymbol{c}, x), f^v(\boldsymbol{c}, x))$.

\subsection{\texttt{LucidAtlas} Formulation}
\label{sec.workflow_lucidatlas}
\subsubsection{Introducing Spatial Dependency}
\label{sec.spatial_dep}
\begin{figure*}
    \centering
    \includegraphics[width=0.85\linewidth]{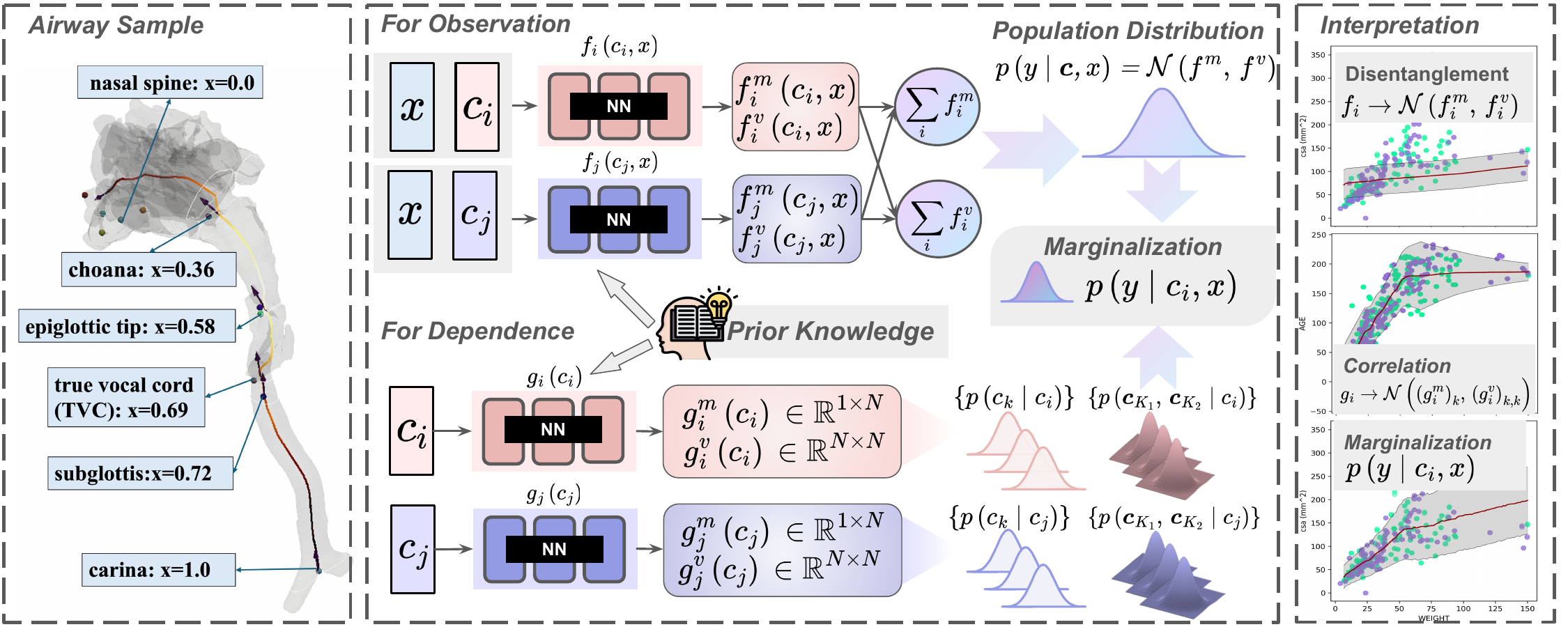}
    \caption{\small \texttt{LucidAtlas}: Learning an Uncertainty-Aware, Covariate-Disentangled, Individualized Atlas Representation. \textcircled{1} As an example use case, we depict an airway with its anatomical landmarks at different depths (i.e., anatomical location) along its centerline~\citep{atlashong2013pediatric}. \textcircled{2} During training, each subnetwork $f_i(c_i, x)$ receives the location $x$ and covariate $c_i$ as input to predict the covariate-specific distributional parameters $f^m_i$ and $f^v_i$, which are added to obtain the overall distributional parameters to capture the population trend and variation as $f^m=\sum_i f^m_i$ and $f^v=\sum_i f^v_i$ respectively. \textcircled{3} The goal of marginalization is to discover $p(y|c_i, x)$, by integrating out the potentially dependent covariates $\{c_k\}_{k \neq i}$. Each subnetwork $g_i(c_i)$ receives covariate $c_i$ to parameterize a multivariate Gaussian distribution $p(\boldsymbol{c}|c_i)$ for all $N$ covariates, from which we obtain $p(c_k|c_i)$ and $p(c_{K_1}, c_{K_2}|c_i)$. The marginalization requires that the outputs of $\{f_i\}$ and $\{g_i\}$ are as described in Sec.~\ref{subsubsec.how_marg}. \textcircled{4} \texttt{LucidAtlas} can obtain different interpretations, i.e., 1) a covariate disentanglement corresponding to  each covariate's additive effect from $\{f_i\}$; 2) dependence between covariates modeled by $\{g_i\}$ as $p(c_k|c_i)$ and 3) a marginalization illustrating the overall impact from each covariate on the predicted response (here the cross-sectional area $y$ at a specific location $x$) via marginalization. \textcircled{5} Monotonic neural networks by construction are  used if the influence of a covariate on the response is assumed to be monotonic based on prior knowledge/domain knowledge; otherwise, a multi-layer perceptron (MLP) is used to parameterize the subnetworks.}
    \label{fig.method}
\end{figure*}

We develop \texttt{LucidAtlas} based on NAMLSS~\citep{thielmann2024namlss}, first extending it by incorporating spatial dependency, which is not explicitly modeled in NAMLSS. To achieve this, we introduce neural subnetworks \(\{f_i(c_i, x)\}\) that predict the distributional parameters of \(p(y|\boldsymbol{c}, x)\). Each network \(f_i(c_i, x)\) has two outputs: \(f^m_i(c_i, x)\) and \(f^v_i(c_i, x)\), which capture the contribution from $c_i$ at location $x$ to the mean and variance of $p(y|\boldsymbol{c},x)$ respectively. The overall population mean and variance are then obtained by summing these individual contributions:
\begin{small}
\begin{equation}
f^m(\boldsymbol{c},x) = \sum_{i=1}^{N} f^m_i(c_i,x) \\,
f^v(\boldsymbol{c},x) = \sum_{i=1}^{N} f^v_i(c_i,x)\,.
\label{eq.follow_namlss}
\end{equation}
\end{small}By explicitly modeling spatial dependencies, \texttt{LucidAtlas} extends NAMLSS to spatial atlas construction.
\paragraph{Loss Function.}
We optimize the subnetworks $\{f_i\}$ by minimizing the negative log-likelihood resulting in the loss function
\begin{small}
\begin{equation}
\mathcal{L}(\{f_i\}, \boldsymbol{c}, x) =  \frac{1}{2}\log(2\pi \cdot f^v(\boldsymbol{c},x)) + \frac{(y - f^m(\boldsymbol{c},x))^2}{2 \cdot f^v(\boldsymbol{c},x)}\,,
\label{eq.nll_loss}
\end{equation}
\end{small} where $y$ is the observation at location $x$ given the  covariates $\boldsymbol{c}$.

\subsubsection{Disentangled Covariate Effects}
\label{sec.dist_cov_effects}
We choose NAMs~\citep{agarwal2020neural,thielmann2024namlss} for atlas representation due to their inherent ability to disentangle covariate effects, enabled by their additive subnetwork design. The disentangled effect of covariate \(c_i\) on the population trend is represented by \(f^m_i(c_i, x)\), while its contribution to the population variation is captured by \(f^v_i(c_i, x)\).

\subsubsection{Prior Knowledge}
\label{sec.prior_knolwedge}
To take a further step, we improve the uncertainty-aware neural additive model~\citep{thielmann2024namlss} by incorporating \emph{prior knowledge} of monotonicity~\citep{kitouni2023mono} for the covariates. Specifically, assuming that the distribution of the response $y$ has a stochastically increasing relationship with respect to a particular covariate $c_j$ (while keeping all other covariates fixed), can be incorporated via the modeling ansatz 
\begin{small}
\begin{equation}
    \frac{\partial {f^m(\boldsymbol{c},x)}}{\partial c_j} = \sum_{i=1}^N \frac{\partial f^m_i(c_i,x)}{\partial c_j}=\frac{\partial f^m_j(c_j,x)}{\partial c_j} \geq 0\,.
\label{eq.prior_know}
\end{equation}
\end{small}
As illustrated in Fig.~\ref{fig.method}, $f_j(c_j)$ can be parameterized using a monotonic Lipschiz neural network by design~\citep{kitouni2023mono} when such prior monotonicity information is available. This design guarantees the monotonicity of \(f_j(c_j)\) by construction and ensures that the interpretations derived from the NAMs align with human prior knowledge. 

\subsection{Rethinking the Neural Additive Model}
\label{subsec.why_mrg}
The underlying assumption behind NAMs is that each covariate contributes independently to the outcome. Even if the covariates are dependent, the NAMs will spread out the contribution from each covariate to its subnetworks as additive functions~\citep{agarwal2020neural}. In most real-world applications, such as our airway atlas construction problem in Sec.~\ref{sec.prob_des}, the independence between covariates cannot be guaranteed. \emph{A natural question is whether accepting the independence assumption unquestioningly and directly using NAMs is appropriate.} In this section, we discuss trustworthiness and the potential risks of neural additive models. 

We can investigate this problem with a toy example, where $y=sin(c_1)+c_2 +\epsilon$ where $\epsilon$ is a noise term and $c_1$,  $c_2$ are covariates that influence the observed outcome $y$. Assuming there is a NAM that already fits this function well, the subnetworks capture $f_1(c_1) = \sin(c_1)$ and $f_2(c_2) = c_2$ and thus approximate $y$ with $f(c_1, c_2) = f_1(c_1) + f_2(c_2)$.

If we want to interpret the population trend of $y$ only with respect to $c_1$, we need to marginalize $c_2$ out as,
\begin{small}
\begin{equation}
\begin{split}
&F_1(c_1)=\int_{-\infty}^{\infty} [f_1(c_1) + f_2(c_2)]p(c_2|c_1) \dif c_2 \\
&=\underbrace{f_1(c_1)}_{\text{Interpretation from NAMs}}+\underbrace{\int_{-\infty}^{\infty} f_2(c_2)p(c_2|c_1) \dif c_2}_{\text{Interpretation from Dependence: } :=h_1(c_1)} 
\end{split}
\label{eq.source_of_intr}
\end{equation}
\end{small}

where $h_1(c_1)$ measures how the dependence between $c_1$ and $c_2$ influences the marginalization $F_1(c_1)$. We can see from Eq.~\eqref{eq.source_of_intr} that $F_1(c_1)$ is composed of the interpretation from the NAM's subnetwork plus the interpretation from the dependence between $c_1$ and $c_2$ as $h_1(c_1)$.

\textbf{If $c_1$ and $c_2$ are \textit{independent}}, $h_1(c_1)=\int_{-\infty}^{\infty} f_2(c_2)p(c_2|c_1) \dif c_2=\int_{-\infty}^{\infty} f_2(c_2)p(c_2) \dif c_2=constant$, which means the marginalization is the actual covariate disentanglement in Eq.~\eqref{eq.follow_namlss} plus a constant. \textbf{If $c_1$ and $c_2$ are \textit{dependent}}, $h_1(c_1)$ is a function of $c_1$ which no longer needs to be a constant and could be a non-trivial function of $c_1$ arising from the inherent stochastic dependence between $c_1$ and $c_2$. Therefore, considering the relationship between $c_1$ and $c_2$ is crucial when using either covariate by itself to interpret the population trend.

In summary, disentangled covariate effects of NAMs, combined with those effects contributed by covariate dependence, shape human-understandable explanations aligned with population trends. \emph{While ignoring potential dependencies in NAMs may not impact prediction performance, it can result in ambiguous or misleading interpretations when analyzing population trends.} More analysis is available in Sec.~\ref{supp.subsec.toy_dataset} in the Supplementary Material.

\subsection{Covariate Marginalization}
\label{subsubsec.how_marg}
\emph{\textcolor{blue}{Due to space constraints, the full derivation is provided in Sec.~\ref{supp.subsubsec.how_marg} of the Supplementary Material.}} This section introduces our proposed marginalization approach to improve the trustworthiness of NAMs when trying to understand the dependency of a covariate on the response.  The dependency of covariates can be modeled with a multivariate Gaussian distribution as $p(\boldsymbol{c}|c_i)=\mathcal{N}(\hat{\boldsymbol{\mu}}(c_i), \hat{\boldsymbol{\Sigma}}(c_i))$ where $\hat{\boldsymbol{\mu}}(c_i)$ represents the mean vector and  $\hat{\boldsymbol{\Sigma}}(c_i)$ the covariance matrix conditioned on $c_i$. From $p(\boldsymbol{c}|c_i)$, one can extract the distribution of an individual covariate $c_k$ conditioned on $c_i$, as $p(c_k|c_i)=\mathcal{N}(\hat{\mu}_k(c_i), \hat{\Sigma}_{k,k}(c_i))$, e.g., how age $c_i$ determines weight $c_k$. One can also extract the joint distribution of $c_{K_1}$ and $c_{K_2}$ conditioned on $c_i$ as $p(c_{K_1}, c_{K_2}|c_i)$ from $p(\boldsymbol{c}|c_i)$, i.e.,
\begin{small}
\begin{equation}
\begin{split}
p(c_{K_1}, c_{K_2}|c_i) =
\mathcal{N}
\left(
\begin{bmatrix}
\hat{\mu}_{K_1}(c_i) \\
\hat{\mu}_{K_2}(c_i) 
\end{bmatrix},
\begin{bmatrix}
\hat{\Sigma}_{K_1, K_1}(c_i) & \hat{\Sigma}_{K_1, K_2}(c_i)  \\
\hat{\Sigma}_{K_2, K_1}(c_i) & \hat{\Sigma}_{K_2, K_2}(c_i)
\end{bmatrix}
\right),
\end{split}
\label{eq.mgd}
\end{equation}
\end{small}
where $\hat{\Sigma}_{K_1, K_2}(c_i)$ is the covariance between $c_{K_1}$ and $c_{K_2}$.

We employ subnetworks $\{g_i(c_i)\}$, each controlled by an individual covariate $c_i$, to model the corresponding conditional distributions $\{p(\boldsymbol{c}|c_i)\}$. Specifically, each subnetwork $g_i(c_i)$ captures a multivariate Gaussian distribution, expressed as: $p(\boldsymbol{c}|c_i) = \mathcal{N}(g^m_i(c_i), g^v_i(c_i))$, where the mean vector $g^m_i(c_i)$ and the covariance matrix $g^v_i(c_i)$ are predicted by \(g_i(c_i)\), as illustrated in Fig.~\ref{fig.method}. 

Next, from Eq.~\eqref{eq.follow_namlss}, the observation $y$ can be formulated as 
\begin{small}
\begin{equation}
y=f^m(\boldsymbol{c},x) + f^v(\boldsymbol{c},x)\cdot \epsilon \\, ~\epsilon \sim \mathcal{N}(0,1) \\.
\end{equation}
\end{small}
With trained $\{f_i\}$ and $\{g_i\}$, we now investigate how $c_i$ influences the distribution of the observation $y$ as $p(y|c_i,x)= \mathcal{N}(\Tilde{\mu}(c_i, x), \Tilde{\sigma}^2(c_i, x))$, where $\Tilde{\mu}(c_i, x)$ is the expectation of $y$ when fixing $c_i$ and $x$, i.e. $\mathrm{E}[y|c_i,x]$; and $\Tilde{\sigma}^2(c_i, x)$ is the variance of $y$ when fixing $c_i$ and $x$, i.e. $\mathrm{Var}(y|c_i,x)$. 

\paragraph{Mean of $p(y|c_i,x)$.}
We expand the two variable case in Sec.~\ref{subsec.why_mrg} to multi-covariates, with the \emph{law of total expectation}~(1)
\begin{small}
\begin{equation}
\begin{split}
&\Tilde{\mu}(c_i, x) = f^m_i(c_i,x) + \int_{-\infty}^{\infty} (\sum_{k \neq i} f^m_{k}(c_{k},x)) p(\boldsymbol{c}_{k \neq i}|c_i) \dif \boldsymbol{c}_{k \neq i} \\
&= f^m_i(c_i,x) + \sum_{k \neq i} \int_{-\infty}^{\infty} f^m_{k}(c_{k},x) p(c_k|c_i) \dif c_k\,, 
\end{split}
\label{eq.E_y_given_ci}
\end{equation}
\end{small} where $f^m_k(c_k)$ represents the interpretation from the additive subnetwork $f_k$ of \texttt{LucidAtlas} and $\boldsymbol{c}_{k \neq i}=[c_1, ..., c_{i-1}, c_{i+1}, ..., c_N]$.

Eq.~\eqref{eq.E_y_given_ci} indicates that even when multiple covariates are involved, only conditional dependencies with respect to individual covariates ($p(c_k|c_i)$) are required to compute $\Tilde{\mu}(c_i, x)$ as a consequence of the additive model for a NAM, which simplifies computations.

\paragraph{Variance of $p(y|c_i,x)$.}
The \emph{law of total variance} is $\mathrm{Var}(Y) =\mathrm{E}[\mathrm{Var}(Y|X)] + \mathrm{Var}(\mathrm{E}[Y|X])$ which states that the total variance of a random variable $Y$ can be broken into two parts: \textcircled{1} the \textbf{expected variance of $Y$ given $X$}, which represents how much $Y$ fluctuates around its mean for each specific value of $X$; and \textcircled{2} The variance of the \textbf{expected value of $Y$ given $X$}, which measures how much the mean of $Y$ changes as $X$ varies.
With the \emph{law of total variance}, 
\begin{small}
\begin{equation}
\mathrm{Var}(y|c_i,x)= \underbrace{\mathrm{E}[\mathrm{Var}(y|\boldsymbol{c}_{k \neq i}, c_i,x)]}_{ \text{\textcircled{1}} :=\Tilde{\sigma}^2_E(c_i, x)} + \underbrace{\mathrm{Var}(\mathrm{E}[y|\boldsymbol{c}_{k \neq i}, c_i,x])}_{\text{\textcircled{2}}:=\Tilde{\sigma}^2_V(c_i, x)}
\label{eq.Var_dist}
\end{equation}
\end{small}
The expected variance of $f^v(\boldsymbol{c}, x)$ given $c_i$ and $x$ is
\begin{small}
\begin{equation}
\begin{split}
&\Tilde{\sigma}^2_E(c_i, x) = f^v_i(c_i,x) + \sum_{k \neq i} \int_{-\infty}^{\infty} f^v_{k}(c_{k},x) p(c_k|c_i) \dif c_k\,.
\end{split}
\label{eq.E_of_V}
\end{equation}
\end{small}
And the variance of the expected value of $f^m(\boldsymbol{c},x)$ given $c_i$ and $x$ can be computed as
\begin{tiny}
\begin{equation}
\begin{split}
&\Tilde{\sigma}^2_V(c_i, x) =\mathrm{Var}(\mathrm{E}[y|\boldsymbol{c}_{k \neq i}, c_i,x])=\mathrm{Var}(\sum_{k \neq i}f^m_k(c_k, x)|c_i, x) \\ &= \sum_{k \neq i} \underbrace{\mathrm{Var}(f^m_k(c_k,x)|c_i, x)}_{\text{\textcircled{3}}} + \mathop{\sum\sum}_{K_1\neq K_2 \neq i} \underbrace{\mathrm{Cov}(f^m_{K_1}(c_{K_1}, x), f^m_{K_2}(c_{K_2}, x) | c_i, x)}_{\text{\textcircled{4}}}
\end{split}
\label{eq.V_of_E}
\end{equation}
\end{tiny}
where
\begin{small}
\begin{equation}
\begin{split}
&\text{\textcircled{3}} = \int_{-\infty}^{\infty} (f^m_k(c_k, x) - \Tilde{\mu}_k(c_i, x))^2p(c_k| c_i) \dif c_k ,\\
& \Tilde{\mu}_k(c_i, x) = \int_{-\infty}^{\infty} f^m_k(c_k, x)p(c_k|c_i)dc_k
\end{split}
\label{eq.V_of_E_part1}
\end{equation}
\end{small}
\begin{tiny}
\begin{equation}
\begin{split}
\text{\textcircled{4}} = \int_{-\infty}^{\infty}\int_{-\infty}^{\infty} f^m_{K_1}(c_{K_1}, x)f^m_{K_2}(c_{K_2}, x)p(c_{K_1}, c_{K_2}|c_i)\dif c_{K_1} \dif c_{K_2} \\
 - \Tilde{\mu}_{K_1}(c_i, x) \Tilde{\mu}_{K_2}(c_i, x) 
\end{split}
\label{eq.V_of_E_part2}
\end{equation}
\end{tiny}


Eqs.~\eqref{eq.E_of_V}-\eqref{eq.V_of_E_part2} imply that instead of sampling the entire covariate space, one only needs to sample from the joint Gaussian distribution between the two covariates, conditioned on the individual covariates, to obtain the marginalized distribution $p(y|c_i,x)$.

\paragraph{Approximation.} 
The integrals in $\Tilde{\mu}(c_i, x)$ (in Eq.~\eqref{eq.E_y_given_ci}), $\Tilde{\sigma}^2_E(c_i, x)$ (in Eq.~\eqref{eq.E_of_V}) and $\Tilde{\sigma}^2_V(c_i, x)$ (in Eqs.~\eqref{eq.E_of_V}-\eqref{eq.V_of_E_part2}) can be approximated using Monte Carlo sampling from $p(c_k|c_i)$ and $p(c_{K_1}, c_{K_2}|c_i)$..

\paragraph{Computational Complexity.}
Suppose there are $N$ covariates and $L$ samples. The computational complexity of marginalizing the NAM for a covariate is \(\mathcal{O}(LN)\), making it feasible in practice. In contrast, for a black-box model, which does not assume our additive structure, the complexity is exponentially higher at \(\mathcal{O}(L^N)\), making direct computation infeasible for large \(N\).

As a result, we obtain $\Tilde{\mu}(c_i, x)$ and $\Tilde{\sigma}^2(c_i, x)=\Tilde{\sigma}^2_E(c_i, x)+\Tilde{\sigma}^2_V(c_i, x)$ to parameterize $p(y|c_i) = \mathcal{N}(\Tilde{\mu}(c_i, x), \Tilde{\sigma}^2(c_i, x))$, capturing the influence of a single covariate $c_i$ on the observation $y$. Our approach aligns with NAM interpretations and can be applied post-hoc.

\subsubsection{Imputation}
\label{sec.imputation}
Our approach naturally facilitates the imputation of missing covariates, as it inherently predicts the conditional distributions \(\{p(c_k \mid c_i)\}\), enabling a principled way to estimate missing values.
 Specifically, if $c_i$ is missing, one can choose the $g_s$ whose uncertainty is the smallest as the predictor for $c_i$ as $s \gets \arg\min_{k,k \neq i} \{{g^v_{k,i}(c_k)}\}$.

\subsubsection{Individualized Prediction}
\label{sec.ind_pred}
One challenge in the context of atlas discovery is to make individualized predictions when observations are predominantly limited to a single time point, i.e., when the atlas is built from cross sectional data. \texttt{LucidAtlas} provides an approach for individualized prediction based on previous observations. Note that this approach is not based on true longitudinal data  (as such data is frequently not available) but instead aims to predict individual future responses based on the cross-sectional population trend.

We define our problem as follows. Given an observation $y^t$ at $x$ with its corresponding covariates $\boldsymbol{c}^t$  at time $t$, how will a subject's response change when $\boldsymbol{c}^t$ changes to $\boldsymbol{c}^{t+1}$ at time $t+1$?
First, we can obtain the probability distribution with  \texttt{LucidAtlas}, as $p(y^t|\boldsymbol{c}^t,x)=\frac{1}{\sqrt{2\pi f^v(\boldsymbol{c}^t,x)}}\exp(-\frac{(y^t-f^m(\boldsymbol{c}^t,x))^2}{2 \cdot f^v(\boldsymbol{c}^t,x)})$.

\begin{assumption}
We assume that the percentile of a subject remains stationary between observations at two nearby time points, i.e., the cumulative distribution, $\mathrm{F}$, should be stationary: $\mathrm{F}(y^t) = \mathrm{F}(y^{t+1})$.
\label{assump1}
\end{assumption}
An intuitive example for Assumption~\ref{assump1} is that if a child has the largest airway among all 2-year-olds, it is likely that this child's airway will remain the largest over a short period of time. Therefore, an approximate individualized prediction can be obtained as $y^{t+1} \approx f^m(\boldsymbol{c}^{t+1},x) + {\sqrt{\frac{f^v(\boldsymbol{c}^{t+1},x)}{f^v(\boldsymbol{c}^{t},x)}}(y^t-f^m(\boldsymbol{c}^t,x))}$\,.

\section{Experiments}
We aim to answer the following questions with our experiments: \textcircled{1} \emph{How well can \texttt{LucidAtlas} estimate population trends?} \textcircled{2} \emph{Can \texttt{LucidAtlas} capture heteroscadastistic variances across a population?} \textcircled{3} \emph{Do explanations from \texttt{LucidAtlas} align with our prior knowledge?} \textcircled{4} \emph{Isaccepting the independence assumption unquestioningly and directly using NAMs  appropriate in scientific discovery?}  \textcircled{5} \emph{How well can \texttt{LucidAtlas} predict responses at  time B given observations at an earlier time A?}

\subsection{Datasets \& Experimental Protocols}

Learning a pediatric airway atlas is the primary motivating problem of our work~\citep{atlashong2013pediatric}. We also use the OASIS brain dataset~\cite{marcus2007oasis} to validate our approach. The Supplementary Material provides more details about these two datasets and our experimental settings in Sec.~\ref{supp.sec.dataset}.

\textbf{Pediatric Airway Geometry.} 
The dataset includes 358 upper airway shapes obtained from computed tomography (CT) images of children with radiographically normal airways. These 358 airway shapes correspond to 264 patients, with 34 having longitudinal observations and 230 who are observed only once. We consider three covariates in this study: age, weight, and height. The majority of the missing data occur in height and the field of view of the airway. Each complete airway has 11 anatomical landmarks, of which 6 are used in our experiments (Fig.~\ref{fig.method}); 263 scans have complete covariate data (age, weight, height).  

We aim to construct an airway atlas which captures airway cross sectional area (CSA) as well as CSA population distributions, incorporating the prior knowledge that CSA shoul monotonically increase with age, weight, and height.  

We convert the $k^{th}$ airway shape into a CSA function mapping normalized depth $x \in [0,1]$ to CSA values as $CSA = C_k(x)$. Based on our discretization complete airways have 500 depth-CSA pairs uniformly distributed on the airway centerline, while incomplete ones have \( \leq 500 \).  

The dataset is split into 80\%/20\% training / test sets respectively by patient, with all longitudinal data in the test set. This ensures that the model learns population trends from individual observations while retaining longitudinal data for individual evaluations.  



\paragraph{OASIS Brain Volumes.} 
Brain segmentations were obtained from the OASIS dataset~\citep{marcus2007oasis}, which includes two subsets:  
\textcircled{1} A cross-sectional set with 416 subjects aged 18–96, primarily single-time observations, plus a reliability subset of 20 non-demented subjects rescanned within 90 days.  
\textcircled{2} A longitudinal set of 150 older adults (60–96 years), totaling 373 imaging sessions.  

Our experiments include four covariates: age, socioeconomic status (SES), mini-mental state examination (MMSE), and clinical dementia rating (CDR). The response variable is the normalized whole brain volume (nWBV).  

\emph{We aim to investigate the relationships between these covariates and brain volume.} Based on prior knowledge, brain volume should not increase with age or CDR, nor decrease when mental state improves~\citep{fotenos2008brain}.


\paragraph{Comparison Methods.}
We choose the current state-of-the-art explainable regression methods, i.e., LightGBM~\citep{ke2017lightgbm} and Explainable Boosting Machines (EBM)~\citep{lou2013EBM} to provide high-quality regression performance. We also compare \texttt{LucidAtlas} with NAM with Exu activations and ensembling strategies~\citep{agarwal2020neural}. We found that NAMLSS performs best for learning population trends, justifying our choice of an additive structure. Besides NAMLSS for uncertainty estimation, we also directly use a multi-layer perceptron (MLP) for mean-variance parameterization of the negative log-likelihood loss as a baseline, termed MLP+NLL. I.e., this MLP+NLL model does not assume an additive structure.


\paragraph{Evaluation Metrics.}
 The Mean Absolute Relative Percent Difference (MARPD) evaluates regression accuracy in capturing population trends, while the Negative Log-Likelihood (NLL) assesses how closely the modeled distribution aligns with the true data distribution. 

\begin{table*}[t]
\centering
\resizebox{0.9\textwidth}{!}{
\begin{tabular}{c|ccc|c|ccccccc}
\hline
 &
   &
   &
   &
  \cellcolor[HTML]{E9FEE9} &
  \multicolumn{7}{c}{\cellcolor[HTML]{E2E3F8}\textbf{Pediatric Airway}} \\ \cline{6-12} 
\multirow{-2}{*}{Methods} &
  \multirow{-2}{*}{Spa.} &
  \multirow{-2}{*}{Add.} &
  \multirow{-2}{*}{Mono.} &
  \multirow{-2}{*}{\cellcolor[HTML]{E9FEE9}\textbf{OASIS Brain}} &
  \multicolumn{1}{c|}{\textbf{Overall}} &
  nasal spine &
  choana &
  epiglottic tip &
  \cellcolor[HTML]{F8E5E4}TVC &
  \cellcolor[HTML]{F8E5E4}subglottis &
  \cellcolor[HTML]{F8E5E4}carina \\ \hline
EBM &
  \XSolidBrush &
  \XSolidBrush &
  \Checkmark &
  3.1920$\pm$2.4985 &
  \multicolumn{1}{c|}{38.8774$\pm$38.9116} &
  44.0259$\pm$40.0036 &
  41.9254$\pm$48.9492 &
  17.7694$\pm$22.4692 &
  20.8076$\pm$26.3942 &
  22.3948$\pm$28.1368 &
  25.4489$\pm$31.4176 \\
LightGBM &
  \XSolidBrush &
  \XSolidBrush &
  \XSolidBrush &
  3.0289$\pm$2.4324 &
  \multicolumn{1}{l|}{36.7755$\pm$35.8704} &
  46.1999$\pm$39.9386 &
  {\color[HTML]{9A0000} \textbf{28.4655$\pm$29.9739}} &
  {\color[HTML]{9A0000} \textbf{14.1479$\pm$12.1693}} &
  19.3283$\pm$16.9426 &
  20.6356$\pm$19.1491 &
  19.9048$\pm$23.6890 \\
NAM &
  \XSolidBrush &
  \Checkmark &
  \XSolidBrush &
  \multicolumn{1}{l|}{3.3882$\pm$2.4748} &
  \multicolumn{1}{l|}{37.1746$\pm$35.1216} &
  \multicolumn{1}{l}{43.8124$\pm$40.1094} &
  \multicolumn{1}{l}{\textbf{37.2806$\pm$33.4382}} &
  \multicolumn{1}{l}{17.4907$\pm$13.0446} &
  \multicolumn{1}{l}{21.0430$\pm$18.8356} &
  \multicolumn{1}{l}{22.5744$\pm$21.0187} &
  \multicolumn{1}{l}{21.8049$\pm$24.0872} \\ \hline
MLP+NLL &
  \XSolidBrush &
  \XSolidBrush &
  \XSolidBrush &
  3.1036$\pm$2.6835 &
  \multicolumn{1}{l|}{38.3364$\pm$36.8091} &
  33.9609$\pm$26.9933 &
  45.0381$\pm$34.6801 &
  18.5780$\pm$17.1130 &
  20.5003$\pm$23.1434 &
  19.9590$\pm$23.7551 &
  21.0536$\pm$13.8406 \\
NAMLSS &
  \XSolidBrush &
  \Checkmark &
  \XSolidBrush &
  {\color[HTML]{000000} \textbf{3.0244$\pm$2.3980}} &
  \multicolumn{1}{l|}{37.3115$\pm$34.7218} &
  {\color[HTML]{9A0000} \textbf{30.7901$\pm$25.6496}} &
  41.1237$\pm$33.3017 &
  {\color[HTML]{000000} 21.5784$\pm$17.1608} &
  {\color[HTML]{000000} 25.3131$\pm$24.5728} &
  {\color[HTML]{000000} 24.2743$\pm$24.6332} &
  {\color[HTML]{000000} 21.1646$\pm$16.1043} \\ \hline
\rowcolor[HTML]{EFEFEF} 
Ours  np &
  \Checkmark &
  \Checkmark &
  \XSolidBrush &
  {\color[HTML]{000000} \textbf{3.0244$\pm$2.3980}} &
  \multicolumn{1}{c|}{\cellcolor[HTML]{EFEFEF}{\color[HTML]{000000} \textbf{36.4472$\pm$35.5650}}} &
  \textbf{32.9029$\pm$25.5350} &
  40.4435$\pm$33.1123 &
  16.1585$\pm$15.4162 &
  {\color[HTML]{9A0000} \textbf{19.2762$\pm$23.1651}} &
  {\color[HTML]{9A0000} \textbf{17.8992$\pm$23.5066}} &
  {\color[HTML]{000000} \textbf{17.4608$\pm$12.2250}} \\
\rowcolor[HTML]{EFEFEF} 
Our part &
  \Checkmark &
  \Checkmark &
  \Checkmark &
  {\color[HTML]{9A0000} \textbf{3.0045$\pm$2.2426}} &
  \multicolumn{1}{c|}{\cellcolor[HTML]{EFEFEF}37.1696$\pm$35.6229} &
  37.4176$\pm$27.0885 &
  42.1177$\pm$33.8561 &
  16.3098$\pm$15.7531 &
  19.7589$\pm$23.1405 &
  19.1237$\pm$23.3031 &
  17.9994$\pm$12.3826 \\
\rowcolor[HTML]{EFEFEF} 
Ours imp &
  \Checkmark &
  \Checkmark &
  \Checkmark &
  3.0936$\pm$2.2473 &
  \multicolumn{1}{c|}{\cellcolor[HTML]{EFEFEF}{\color[HTML]{9A0000} \textbf{36.2685$\pm$35.6709}}} &
  34.1421$\pm$26.2881 &
  42.3398$\pm$33.8903 &
  \textbf{15.2841$\pm$14.7377} &
  \textbf{19.2930$\pm$22.9405} &
  \textbf{18.6742$\pm$23.1107} &
  {\color[HTML]{9A0000} \textbf{17.3280$\pm$12.0628}} \\ \hline
\end{tabular}}
\caption{\small Quantitative Evaluation of Normalized Brain Volume Regression (OASIS Brain Dataset) and Cross-Sectional Area Regression (Pediatric Airway Dataset) with respect to Mean Absolute Relative Percent Difference (MARPD, \%). We also evaluate with respect to  different landmarks. The \{TVC, subglottic and carina\} landmarks are significant landmarks for airway obstruction analysis~\citep{atlashong2013pediatric}. \textbf{\textcolor{purple}{Bold red values}} indicate the best scores across all methods. \textbf{Bold black values} indicate the 2nd best scores of all methods. \textbf{Spa.} indicates whether spatial dependency is considered. \textbf{Add.} indicates whether a model has an additive design. \textbf{Mono.} indicates whether prior knowledge about monotonicity is used. \textit{Ours np} refers to \texttt{LucidAtlas} without incorporating prior knowledge. \textit{Ours part} denotes our model trained only on complete data, while \textit{Ours imp} represents using the full dataset for training, including missing values. \texttt{LucidAtlas} performs best overall.}
\label{exp.single_quant_marpd}
\end{table*}

\begin{table}[t]
\centering
\resizebox{0.5\textwidth}{!}{
\begin{tabular}{c|ccc|c|ccccccc}
\hline
\multicolumn{1}{l|}{} &
   &
   &
   &
  \cellcolor[HTML]{E9FEE9} &
  \multicolumn{7}{c}{\cellcolor[HTML]{E2E3F8}\textbf{Pediatric Airway}} \\ \cline{6-12} 
\multicolumn{1}{l|}{\multirow{-2}{*}{}} &
  \multirow{-2}{*}{Spa.} &
  \multirow{-2}{*}{Add.} &
  \multirow{-2}{*}{Mono.} &
  \multirow{-2}{*}{\cellcolor[HTML]{E9FEE9}\textbf{OASIS Brain}} &
  \multicolumn{1}{c|}{\textbf{Overall}} &
  nasal spine &
  choana &
  epiglottic tip &
  \cellcolor[HTML]{F8E5E4}TVC &
  \cellcolor[HTML]{F8E5E4}subglottis &
  \cellcolor[HTML]{F8E5E4}carina \\ \hline
MLP+NLL &
  \XSolidBrush &
  \XSolidBrush &
  \XSolidBrush &
  {\color[HTML]{000000} 2.5573} &
  \multicolumn{1}{c|}{{\color[HTML]{000000} 1.0273}} &
  {\color[HTML]{9A0000} \textbf{1.7403}} &
  {\color[HTML]{000000} 1.7702} &
  {\color[HTML]{000000} -0.1508} &
  {\color[HTML]{000000} -0.0302} &
  {\color[HTML]{000000} 0.0778} &
  {\color[HTML]{000000} 0.2415} \\
NAMLSS &
  \XSolidBrush &
  \Checkmark &
  \XSolidBrush &
  {\color[HTML]{000000} 0.6714} &
  \multicolumn{1}{c|}{{\color[HTML]{000000} 1.0020}} &
  {\color[HTML]{000000} \textbf{1.7591}} &
  {\color[HTML]{9A0000} \textbf{1.4011}} &
  {\color[HTML]{000000} -0.0561} &
  {\color[HTML]{000000} 0.4015} &
  {\color[HTML]{000000} 0.3951} &
  {\color[HTML]{000000} 0.917} \\ \hline
\rowcolor[HTML]{EFEFEF} 
Ours  np &
  \Checkmark &
  \Checkmark &
  \XSolidBrush &
  {\color[HTML]{000000} 0.6714} &
  \multicolumn{1}{c|}{\cellcolor[HTML]{EFEFEF}{\color[HTML]{000000} 0.9365}} &
  {\color[HTML]{000000} 1.7796} &
  {\color[HTML]{000000} \textbf{1.4115}} &
  {\color[HTML]{000000} \textbf{-0.3948}} &
  {\color[HTML]{000000} \textbf{-0.1274}} &
  {\color[HTML]{9A0000} \textbf{-0.1633}} &
  {\color[HTML]{9A0000} \textbf{0.0331}} \\
\rowcolor[HTML]{EFEFEF} 
Our part &
  \Checkmark &
  \Checkmark &
  \Checkmark &
  {\color[HTML]{9A0000} \textbf{0.6457}} &
  \multicolumn{1}{c|}{\cellcolor[HTML]{EFEFEF}{\color[HTML]{9A0000} \textbf{0.8467}}} &
  1.9669 &
  1.4561 &
  -0.3043 &
  {\color[HTML]{9A0000} \textbf{-0.1293}} &
  \textbf{-0.1185} &
  \textbf{0.0726} \\
\rowcolor[HTML]{EFEFEF} 
Ours imp &
  \Checkmark &
  \Checkmark &
  \Checkmark &
  \textbf{0.6514} &
  \multicolumn{1}{c|}{\cellcolor[HTML]{EFEFEF}\textbf{0.8901}} &
  1.8087 &
  1.4542 &
  {\color[HTML]{9A0000} \textbf{-0.4341}} &
  -0.0065 &
  0.0146 &
  0.0966 \\ \hline
\end{tabular}
}
\caption{\small Quantitative Evaluation of Population Distribution Estimation based on Negative Log-Likelihood (NLL). Our approach achieves the best performance overall.  }
\label{exp.single_quant_nll}
\end{table}

\begin{table}[t]
\centering
\resizebox{0.5\textwidth}{!}{
\begin{tabular}{ccccccccc}
\hline
\rowcolor[HTML]{E2E3F8} 
\multicolumn{9}{c}{\cellcolor[HTML]{E2E3F8}\textbf{Pediatric Airway}} \\ \hline
\multicolumn{1}{c|}{Covariate} &
  \multicolumn{1}{c|}{Corr.} &
  \multicolumn{1}{c|}{Overall} &
  nasal spine &
  choana &
  epiglottic tip &
  \cellcolor[HTML]{F8E5E4}TVC &
  \cellcolor[HTML]{F8E5E4}subglottis &
  \cellcolor[HTML]{F8E5E4}carina \\ \hline
\multicolumn{1}{c|}{Age} &
  \multicolumn{1}{c|}{\XSolidBrush} &
  \multicolumn{1}{c|}{0.9907} &
  {\color[HTML]{9A0000} \textbf{1.9341}} &
  1.4836 &
  0.4278 &
  {\color[HTML]{000000} 0.1805} &
  0.1673 &
  0.5516 \\
\rowcolor[HTML]{EFEFEF} 
\multicolumn{1}{c|}{\cellcolor[HTML]{EFEFEF}Age} &
  \multicolumn{1}{c|}{\cellcolor[HTML]{EFEFEF}\Checkmark} &
  \multicolumn{1}{c|}{\cellcolor[HTML]{EFEFEF}{\color[HTML]{9A0000} \textbf{0.8565}}} &
  {\color[HTML]{000000} 2.0124} &
  {\color[HTML]{9A0000} \textbf{1.4789}} &
  {\color[HTML]{9A0000} \textbf{-0.183}} &
  {\color[HTML]{9A0000} \textbf{-0.1709}} &
  {\color[HTML]{9A0000} \textbf{-0.1777}} &
  {\color[HTML]{9A0000} \textbf{0.1485}} \\ \hline
\multicolumn{1}{c|}{Height} &
  \multicolumn{1}{c|}{\XSolidBrush} &
  \multicolumn{1}{c|}{0.9830} &
  2.1449 &
  1.4496 &
  0.4049 &
  0.2185 &
  0.2179 &
  0.4715 \\
\rowcolor[HTML]{EFEFEF} 
\multicolumn{1}{c|}{\cellcolor[HTML]{EFEFEF}Height} &
  \multicolumn{1}{c|}{\cellcolor[HTML]{EFEFEF}\Checkmark} &
  \multicolumn{1}{c|}{\cellcolor[HTML]{EFEFEF}{\color[HTML]{9A0000} \textbf{0.8330}}} &
  {\color[HTML]{9A0000} \textbf{1.9187}} &
  {\color[HTML]{9A0000} \textbf{1.3968}} &
  {\color[HTML]{9A0000} \textbf{-0.2574}} &
  {\color[HTML]{9A0000} \textbf{-0.1002}} &
  {\color[HTML]{9A0000} \textbf{-0.0819}} &
  {\color[HTML]{9A0000} \textbf{0.0968}} \\ \hline
\multicolumn{1}{c|}{Weight} &
  \multicolumn{1}{c|}{\XSolidBrush} &
  \multicolumn{1}{c|}{1.0804} &
  2.1305 &
  {\color[HTML]{9A0000} \textbf{1.3957}} &
  0.6239 &
  0.4729 &
  0.4701 &
  0.7822 \\
\rowcolor[HTML]{EFEFEF} 
\multicolumn{1}{c|}{\cellcolor[HTML]{EFEFEF}Weight} &
  \multicolumn{1}{c|}{\cellcolor[HTML]{EFEFEF}\Checkmark} &
  \multicolumn{1}{c|}{\cellcolor[HTML]{EFEFEF}{\color[HTML]{9A0000} \textbf{0.8813}}} &
  {\color[HTML]{9A0000} \textbf{1.9122}} &
  1.431 &
  {\color[HTML]{9A0000} \textbf{-0.1077}} &
  {\color[HTML]{9A0000} \textbf{0.0004}} &
  {\color[HTML]{9A0000} \textbf{0.0237}} &
  {\color[HTML]{9A0000} \textbf{0.1829}} \\ \hline
\end{tabular}
}
\caption{\small Quantitative Comparison of Different Ways of Marginalization. NLL is computed between the marginalized covariate interpretation and the data distribution. A \Checkmark in the \textbf{Corr.} column indicates that covariate dependence is considered, while \XSolidBrush signifies that it is ignored. Accounting for covariate dependence improves alignment between covariate interpretation and the data distribution.
}
\label{exp.airway_cov_corr}
\end{table}

\begin{table}[t]
\centering
\resizebox{0.5\textwidth}{!}{
\begin{tabular}{l|l|llllllllllllll}
\hline
\multicolumn{1}{c|}{} &
  \multicolumn{1}{c|}{\cellcolor[HTML]{E9FEE9}} &
  \multicolumn{14}{c}{\cellcolor[HTML]{E2E3F8}\textbf{Pediatric Airway}} \\ \cline{3-16} 
\multicolumn{1}{c|}{\multirow{-2}{*}{Time}} &
  \multicolumn{1}{c|}{\multirow{-2}{*}{\cellcolor[HTML]{E9FEE9}\textbf{OASIS Brain}}} &
  \multicolumn{1}{c|}{\textbf{Overall}} &
  \multicolumn{2}{c}{nasal spine} &
  \multicolumn{2}{c}{choana} &
  \multicolumn{2}{c}{epiglottic tip} &
  \multicolumn{3}{c}{\cellcolor[HTML]{F8E5E4}TVC} &
  \multicolumn{2}{c}{\cellcolor[HTML]{F8E5E4}subglottis} &
  \multicolumn{2}{c}{\cellcolor[HTML]{F8E5E4}carina} \\ \hline
T0 &
  1.6042 &
  \multicolumn{1}{l|}{37.2944} &
  \multicolumn{2}{l}{31.2867} &
  \multicolumn{2}{l}{50.55989} &
  \multicolumn{2}{l}{11.9559} &
  \multicolumn{3}{l}{13.2770} &
  \multicolumn{2}{l}{14.2701} &
  \multicolumn{2}{l}{19.1819} \\
Pop. &
  3.1563 &
  \multicolumn{1}{l|}{37.8092} &
  \multicolumn{2}{l}{39.1704} &
  \multicolumn{2}{l}{{\color[HTML]{9A0000} \textbf{45.2991}}} &
  \multicolumn{2}{l}{14.7017} &
  \multicolumn{3}{l}{15.2413} &
  \multicolumn{2}{l}{15.2040} &
  \multicolumn{2}{l}{{\color[HTML]{9A0000} \textbf{17.1568}}} \\ \hline
\rowcolor[HTML]{EFEFEF} 
Ind. &
  {\color[HTML]{9A0000} \textbf{1.4687}} &
  \multicolumn{1}{l|}{\cellcolor[HTML]{EFEFEF}{\color[HTML]{9A0000} \textbf{35.7221}}} &
  \multicolumn{2}{l}{\cellcolor[HTML]{EFEFEF}{\color[HTML]{9A0000} \textbf{31.1910}}} &
  \multicolumn{2}{l}{\cellcolor[HTML]{EFEFEF}50.1397} &
  \multicolumn{2}{l}{\cellcolor[HTML]{EFEFEF}{\color[HTML]{9A0000} \textbf{9.2835}}} &
  \multicolumn{3}{l}{\cellcolor[HTML]{EFEFEF}{\color[HTML]{9A0000} \textbf{10.2199}}} &
  \multicolumn{2}{l}{\cellcolor[HTML]{EFEFEF}{\color[HTML]{9A0000} \textbf{11.2128}}} &
  \multicolumn{2}{l}{\cellcolor[HTML]{EFEFEF}17.2563} \\ \hline
\end{tabular}
}
\caption{\small Mean Absolute Relative Percent Difference (in \%) for Individualized Prediction. \textbf{T0} in the \textbf{Time} column indicates directly using the observation from the initial time point $T0$ to predict at time $T1$. \textbf{Pop.} indicates ignoring the observation at $T0$ and directly using the mean population value $f^m(\boldsymbol{c},x)$ for individualized prediction for $T1$. \textbf{Ind.} indicates our approach illustrated in Sec.~\ref{sec.ind_pred}. Individualized prediction provides the best performance for both datasets and for most landmarks. 
 }
\label{exp.quant_long_marpd}
\end{table}

\begin{figure}[t]
    \centering
    \includegraphics[width=0.95\linewidth]{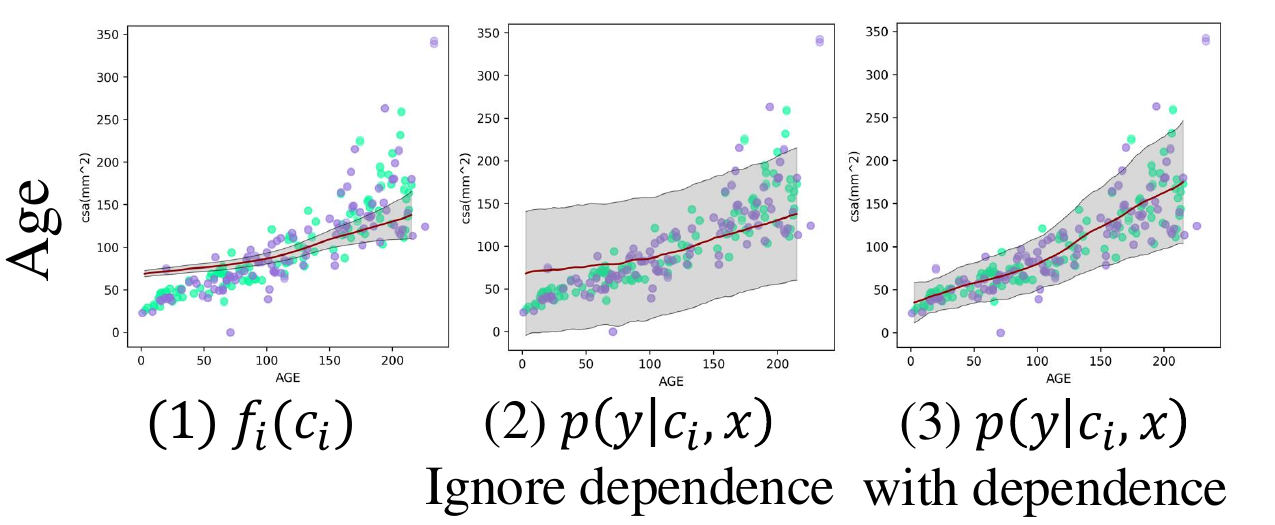}
    \caption{\small Visualizations of Covariate Interpretations from \texttt{LucidAtlas} for CSA Distribution at the Subglottis Landmark (Pediatric Airway Dataset). (1) $f_i(c_i)$ represents the disentangled covariate effect directly from a NAM as illustrated in Sec.~\ref{sec.dist_cov_effects}; (2) Marginalized covariate interpretation without accounting for covariate dependence; (3) Marginalized covariate interpretation incorporating covariate dependence. \textcolor{SeaGreen}{Green} and \textcolor{Orchid}{purple} dots indicate training and testing samples respectively. The \textcolor{BrickRed}{red} lines represent the learned population trend, and the \textcolor{Gray}{gray} shading spans \(\pm 2 \times\) standard deviations. Considering covariate dependence is essential for accurately capturing how each covariate influences the population trend and associated uncertainties.} 
    \label{fig.vis_whether_do_correlation}
\end{figure}

\begin{figure}[t]
    \centering
    \includegraphics[width=0.97\linewidth]{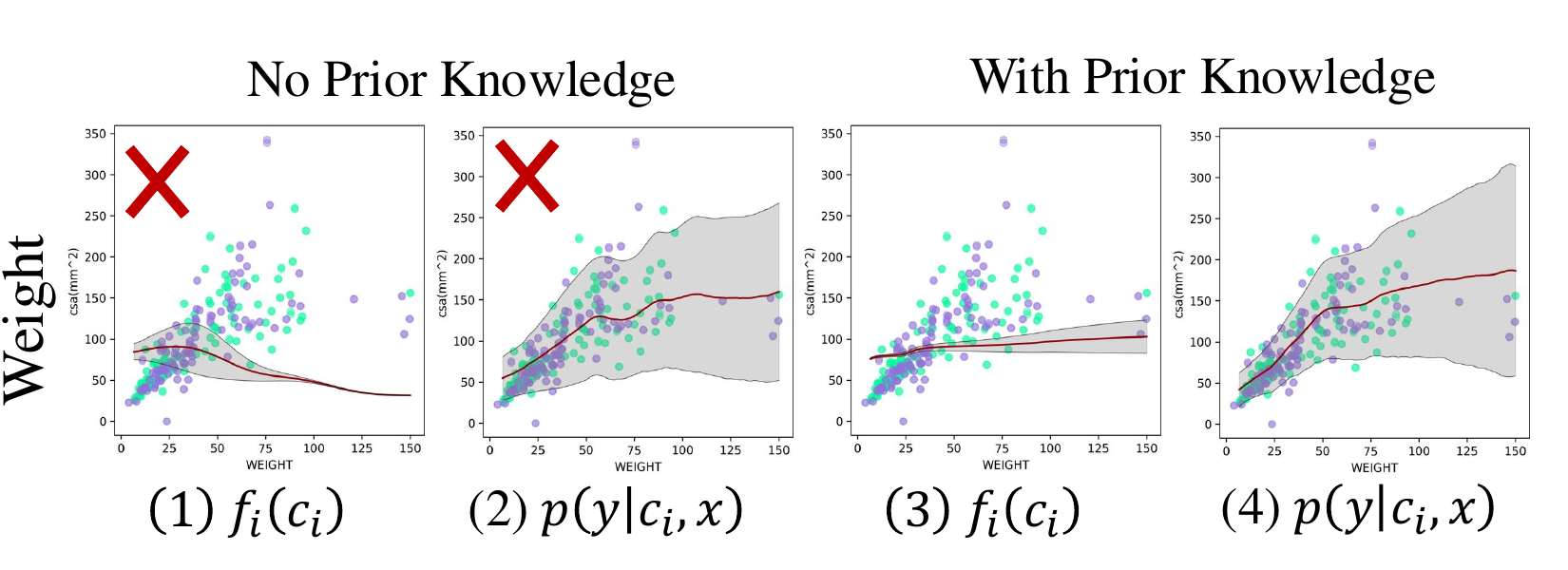}
    \caption{\small Visualizations of the Effect of Prior Knowledge in \texttt{LucidAtlas} at the Subglottis Landmark (Pediatric Airway Dataset). The \textcolor{red}{$\times$} symbol indicates the covariate interpretation contradicts prior knowledge, such as the NAM incorrectly interpreting airway CSA as decreasing with a child's weight. Without incorporating prior knowledge, the model may deviate form our prior assumptions. Without marginalization, to account for covariate dependencies, the data may not be fit well.}
    \label{fig.whether_use_prior}
\end{figure}


\subsection{Discussions}
\textbf{Population Trend Regression.} Table~\ref{exp.single_quant_marpd} presents the quantitative evaluation of population trend regression. Notably, in the absence of a spatial variable $x$, such as in the OASIS Brain dataset, \texttt{LucidAtlas} is equivalent to NAMLSS when no prior knowledge is incorporated. NAMLSS outperforms \textit{MLP+NLL}, highlighting the advantage of an additive network structure with uncertainty estimation in capturing population trends. Among all methods, \texttt{LucidAtlas} achieves the best performance. For the pediatric airway dataset, training with data that include missing values, as described in Sec.~\ref{sec.imputation}, further enhances regression accuracy. 
We observe that the standard deviations of the MARPD metrics in Table~\ref{exp.single_quant_marpd} are large, largely due to data uncertainties. Beyond regression performance, our primary concern is how well the distribution captured by \texttt{LucidAtlas} aligns with the true population distribution, as this accounts for data uncertainties.
\textbf{Population Distribution.} Table~\ref{exp.single_quant_nll} quantifies the performance of different approaches in estimating population distributions. Our approach outperforms all baselines, demonstrating that incorporating prior knowledge of monotonicity, as implemented in \texttt{LucidAtlas}, enhances NAMs' capability in modeling population distributions.
\textbf{Benefit of Prior Knowledge.} Fig.~\ref{fig.whether_use_prior} compares covariate interpretations with and without prior knowledge in \texttt{LucidAtlas} at the subglottis landmark. While the marginalized covariate interpretations in Fig.~\ref{fig.whether_use_prior}(2, 4) may appear similar, disregarding prior knowledge leads to unreasonable covariate interpretations. For instance, the model erroneously interprets airway CSA as decreasing with respect to a child’s weight at the subglottis landmark. Additional visualizations are provided in the Supplementary Material.
\textbf{Role of Covariate Dependence.}
Table~\ref{exp.airway_cov_corr} and Fig.~\ref{fig.vis_whether_do_correlation} examine the significance of covariate dependence in single-feature interpretation. Fig.~\ref{fig.vis_whether_do_correlation} illustrates that ignoring covariate dependence results in underconfident uncertainty estimation and suboptimal population trend prediction, highlighting the necessity of incorporating covariate dependence for reliable interpretation. The quantitative results in Table~\ref{exp.airway_cov_corr} further confirm that, marginalized covariate interpretation with dependence better aligns with the data distribution.
\textbf{Individualized Prediction.} We quantify the performance of individualized predictions (to predict for $T1$ given the observation at $T0$). As described in Sec.~\ref{sec.ind_pred}, our approach, which intergrates the observations at $T0$ with population trends, yields the best predictive performance. More experiment results and visualizations are available at Sec.~\ref{supp.more_vis_exp}.

\section{Limitations and Future Work}
\label{sec.limit}
\texttt{LucidAtlas} models population variance with a Gaussian distribution. Expanding beyond Gaussian assumptions, more flexible probabilistic frameworks—such as non-parametric approaches or mixture models—could improve expressiveness and model fits.
Identifiability issues arise when covariates are dependent or the latent space is redundant, potentially affecting interpretability~\citep{zhou2020identifiability, siems2023curve}. Addressing these concerns is crucial for ensuring well-posed solutions.
Another key extension is incorporating categorical variables into uncertainty quantification, as such variables often exist for real-world datasets. 
Currently, we describe the airway using cross-sectional area. In future work, we plan to develop a probabilistic representation for 3D shape modeling with uncertainties by extending NAISR~\citep{jiao2023naisr}.

\section{Conclusions}
We introduced \texttt{LucidAtlas}, an approach for learning an uncertainty-aware, covariate-disentangled, and individualized atlas representation. Additionally, we highlighted potential risks in using NAMs for covariate interpretation in the presence of covariate dependence and proposed a computationally efficient marginalization approach to mitigate these limitations. Furthermore, we found that incorporating prior knowledge helps eliminate misleading interpretations. We evaluated our method using two distinct datasets, validating its trustworthiness and effectiveness. \texttt{LucidAtlas} stands out as the only atlas representation capable of addressing covariate-, subject-, population-, and spatial-level questions in an interpretable and reliable manner. 

\section{Acknowledgement}
The research reported in this publication was supported by NIH grant 1R01HL154429 and 1R21HL172230-01A1. The content is solely the responsibility of the authors and does not necessarily represent the official views of the NIH. 

\newpage
\clearpage

\bibliography{uai2025-template}

\clearpage
\newpage
\appendix
\input{supplementary}
\onecolumn
\end{document}

%% file: Supplementary.tex
\onecolumn
\renewcommand{\theequation}{S.\arabic{equation}}
\renewcommand{\thefigure}{S.\arabic{figure}}
\renewcommand{\thetable}{S.\arabic{table}}
\renewcommand{\thesection}{S.\arabic{section}}
\section*{\centering{Supplementary Material for \texttt{LucidAtlas}}}

\section{Extended Related Work}
\label{supp.sec.supp_related_work}
\paragraph{Epistemic Uncertainty versus Aleatoric Uncertainty.}
Epistemic and aleatoric uncertainties are two different kinds of uncertainties. Epistemic uncertainty relates to model parameters and stems from limited model knowledge, which is reducible with more data or better modeling. Important techniques include the Laplace approximation~\citep{daxberger2021laplace}, Ensembling~\citep{hullermeier2021survey} and MC-Dropout~\citep{gal2016dropout}. Aleatoric uncertainty arises from inherent data randomness and is irreducible. Important techniques includes a line of Bayesian Neural Networks~\citep{stirn2022faithfulheteroscedasticregressionneural, immer2024effectiveheterbayes}. DeepEnsembles~\citep{lakshminarayanan2017deepemsemble} can handle both epistemic and aleatoric uncertainties.

Regarding uncertainty estimation for interpretable models, more attention is paid to epistemic uncertainties. NAMs used ensembling to estimate and decrease model uncertainties~\citep{agarwal2020neural}. LA-NAM used Laplace approximations for uncertainty estimation~\citep{bouchiat2023lanam} with NAMs. In atlas construction, aleatoric uncertainty is especially important when individual differences are large. Capturing aleatoric uncertainty is crucial in medicine to understand population variations. NAMLSS can model aleatoric uncertainty using NAMs to approximate the parameters $\{\theta^{k}\}$ of a data distribution ~\citep{thielmann2024namlss}, as
\begin{small}
\begin{equation}
\theta^{(k)}=h^{(k)}\left(\beta^{(k)}+\sum_{i=1}^{N} f_{i}^{(k)}\left(c_{i}\right)\right) 
\label{supp.eq.namlss}
\end{equation}
\end{small}
where $\theta^{(k)}$ can, for example, be the mean and variance of Gaussian distributions; $\beta^{(k)}$ denotes the parameter-specific intercept and $f_{i}^{(k)}$ represents the feature network for parameter $k$ for the $i$-th feature. \emph{\texttt{LucidAtlas} extends NAMLSS to a more versatile representation, enabling individualized prediction, incorporating prior knowledge, and capturing spatial dependence.}

\paragraph{Monotonicity.}
Monotonic neural networks ensure that a network's output changes monotonically with respect to certain inputs. Research has focused on two lines of approaches:  architectures such as Deep Lattice Networks~\citep{you2017deep} that guarantee monotonicity but may lack expressiveness, and heuristic methods such as Certified Monotonic Neural Networks \citep{liu2020certified} that use regularization but can be computationally expensive. Recent advancements, including Constrained Monotonic Neural Networks \citep{runje2023constrained}, aim to balance monotonicity, expressiveness, and efficiency. Additionally, research in normalizing flows~\citep{de2020block} has contributed to developing monotonic functions in neural networks to ensure invertibility. Expressive monotonic neural networks~\citep{kitouni2023mono} are constructed using Lipschitz-constrained neural networks, ensuring monotonicity by design while preserving expressiveness.
\emph{We use the Lpipshitz-contrained neural networks to ensure monotonicity in \texttt{LucidAtlas} to follow prior / domain knowledge.}

\paragraph{Disentangled Representation Learning.}
Disentangled representation learning (DRL) has been explored in a variety of domains, including computer vision~\citep{shoshan2021gan, ding2020guided, zhang2018unsupervised, zhang2018learning, xu2021learning, yang2020dsm}, natural language processing~\citep{john2018disentangled}, and medical image analysis~\citep{chartsias2019disentangled, bercea2022federated}. 

Medical data is typically associated with various covariates which should be taken into account during analyses. Taking~\citep{chu2022disentangled} as an example, when observing a tumor's progression, it is difficult to know whether the variation of a tumor's progression is due to time-varying covariates or due to treatment effects. Therefore, being able to disentangle different effects is highly useful for a representation to promote understanding and to be able to quantify the effect of covariates on observations. \emph{\texttt{LucidAtlas} disentangles covariate effects in terms of their contribution to population trends and uncertainties.}

\paragraph{Explainable Artificial Intelligence.}
The goal of eXplainable Artificial Intelligence (XAI) is to provide human-understandable explanations for the decisions and actions of AI models. Various approaches to XAI have been proposed, including counterfactual inference~\citep{berrevoets2021learning, moraffah2020causal, thiagarajan2020calibrating, chen2022covariate}, attention maps~\citep{zhou2016cvpr, jung2021towards, woo2018cbam}, feature importance~\citep{arik2021tabnet, ribeiro2016should, agarwal2020neural}, and instance retrieval~\citep{Crabbe2021Simplex}. A neural additive model (NAM) \citep{agarwal2020neural, jiao2023naisr} is an important XAI method that achieves interpretability through a linear combination of neural networks, each focusing on a \emph{single} input feature. 
NAISR pioneers the use of NAMs for modeling medical shapes to enable scientific discoveries in the medical domain ~\citep{jiao2023naisr}; however, it does not account for heteroskedasticity in its shape representation and does not consider uncertainties. \emph{\texttt{LucidAtlas} extends this concept by integrating NAMs to construct an atlas that captures population trends and uncertainties with spatial dependencies}.

\section{Method}

\subsection{Notations}
Table~\ref{supp.tab.exp_notations} shows the notations used in this paper. 
\begin{table}[]
\resizebox{0.9\textwidth}{!}{
\begin{tabular}{l|l}
\hline
Notations                         & Explanations                                                                          \\ \hline
$y$                               & Observed variable, i.e., target variable to model                                     \\ \hline
$\boldsymbol{c}$                  & A vector containing all $N$ covariates, e.g, $\boldsymbol{c}=[age, weight,...]$       \\ \hline
$f^m(\boldsymbol{c}, x)$ or $f^m$ & Prediction of mean population trend given $\boldsymbol{c}$ at location $x$            \\ \hline
$f^m_i(c_i, x)$ or $f^m_i$       & Additive effects predicted from $i^{th}$ subsnetwork $f_i$ for mean \\ \hline
$f^v(\boldsymbol{c}, x)$ or $f^v$ & Prediction of population variance given $\boldsymbol{c}$ at location $x$              \\ \hline
$f^v_i(c_i, x)$ or $f^v_i$ & \begin{tabular}[c]{@{}l@{}}Additive effects predicted from $i^{th}$ subsnetwork $f_i$ for variance\end{tabular} \\ \hline
$g^m_{i,k}(c_i)$                  & The predicted mean of $c_k$ given $c_i$                                               \\ \hline
$g^v_{i,k}(c_i)$                  & The predicted variance of $c_k$ given $c_i$                                           \\ \hline
$p(y|c_i,x)$                      & Marginalized covariate effects: how $c_i$ affect $y$ at location $x$                  \\ \hline
$\mathrm{E}[y|c_i,x]$             & The expectation of $y$ when $c_i$ and $x$ are fixed                                   \\ \hline
$\mathrm{Var}(y|c_i,x)$           & The variance of $y$ when $c_i$ and $x$ are fixed                                      \\ \hline
\end{tabular}}
\caption{Illustrations of the Notations.}
\label{supp.tab.exp_notations}
\end{table}

\subsection{Expanded Discussion on the Toy Example}
\label{supp.subsec.toy_dataset}
Assuming $c_1$ and $c_2$ are covariates that influence the observed result $y$, a NAM fits well whose subnetworks capture $f_1(c_1) = \sin(c_1)$ and $f_2(c_2) = c_2$ and thus approximate $y$ with $y \approx f(c_1, c_2) + \epsilon = f_1(c_1) + f_2(c_2) + \epsilon$, where $\epsilon$ is  Gaussian noise with mean zero.

If we want to interpret the population trend of $y$ with only $c_1$, we need to marginalize $c_2$ out as
\begin{small}
\begin{equation}
\begin{split}
F_1(c_1)&=\int_{-\infty}^{\infty} [f_1(c_1) + f(c_2)]p(c_2|c_1) \dif c_2 \\
&=\underbrace{f_1(c_1)}_{\text{Interpretation from NAMs}}+\underbrace{\int_{-\infty}^{\infty} f_2(c_2)p(c_2|c_1) \dif c_2}_{\text{Interpretation from Dependence: } :=h_1(c_1)} \\
&= \sin(c_1) + \int_{-\infty}^{\infty} c_2p(c_2|c_1) \dif c_2
\end{split}
\label{supp.eq.source_of_intr1}
\end{equation}
\end{small}

where $h_1(c_1)$ measures how the dependence between $c_1$ and $c_2$ influences the marginalization $F_1(c_1)$. We can see from Eq.~\ref{supp.eq.source_of_intr1} that $F_1(c_1)$ is composed of the interpretation from the NAM's subnetwork plus the interpretation from the dependence between $c_1$ and $c_2$ as $h_1(c_1)$.

If we want to interpret the population trend of $y$ with only $c_2$, we need to marginalize $c_1$ out as
\begin{small}
\begin{equation}
\begin{split}
F_2(c_2)&=\int_{-\infty}^{\infty} [f_2(c_2) + f(c_1)]p(c_1|c_2) \dif c_1 \\
&=\underbrace{f_2(c_2)}_{\text{Interpretation from NAMs}}+\underbrace{\int_{-\infty}^{\infty} f_1(c_1)p(c_1|c_2) \dif c_1}_{\text{Interpretation from Dependence: } :=h_2(c_2)} \\
&= c_2 + \int_{-\infty}^{\infty} \sin(c_1)p(c_1|c_2) \dif c_1\,.
\end{split}
\label{supp.eq.source_of_intr2}
\end{equation}
\end{small}

\textbf{If $c_1$ and $c_2$ are \textit{independent}}, \\

\begin{small}
\begin{equation}
\begin{split}
h_1(c_1)=\int_{-\infty}^{\infty} f_2(c_2)p(c_2|c_1) \dif c_2 =\int_{-\infty}^{\infty} f_2(c_2)p(c_2) \dif c_2=\mathrm{E}_{p(c_2)}[f_2(c_2)]= f_2(\mathrm{E}[c_2])=\mathrm{E}[c_2]=constant\,, 
\end{split}
\label{supp.eq.h1}
\end{equation}
\end{small}

\begin{small}
\begin{equation}
\begin{split}
h_2(c_2)=\int_{-\infty}^{\infty} f_1(c_1)p(c_1|c_2) \dif c_1 =\int_{-\infty}^{\infty} f_1(c_1)p(c_1) \dif c_1=\mathrm{E}_{p(c_1)}[f_1(c_1)]= f_1(\mathrm{E}[c_1])=\sin(\mathrm{E}[c_1])=constant\,. 
\end{split}
\label{supp.eq.h2}
\end{equation}
\end{small}

Thus

\begin{small}
\begin{equation}
\begin{split}
&F_1(c_1)=\sin(c_1) + \mathrm{E}[c_2] \\
&F_2(c_2)=c_2 + \sin(\mathrm{E}[c_1]) 
\end{split}
\label{supp.eq.F1F2}
\end{equation}
\end{small}

which means the marginalization is the actual covariate disentanglement in Sec.~\ref{sec.dist_cov_effects} plus a constant. \\

\textbf{If $c_1$ and $c_2$ are \textit{dependent}} 

$h(c_1)$ is a function of $c_1$ which is controlled by the dependence between $c_1$ and $c_2$.

For example, assume the relationship between  $c_1$ and $c_2$ are at one extreme of dependence in the sense that $c_2 $ is a deterministic function of $c_1$  as 

\begin{small}
\begin{equation}
c_2 = \exp(c_1)\,.
\label{supp.eq.exp_c1}
\end{equation}
\end{small}

Then
\begin{small}
\begin{equation}
\begin{split}
F_1(c_1) &= \sin(c_1) + \int_{-\infty}^{\infty} c_2p(c_2|c_1) \dif c_2 \\
&=\sin(c_1) + \exp(c_1)
\end{split}
\label{supp.eq.F1_with_corr}
\end{equation}
\end{small}

\begin{small}
\begin{equation}
\begin{split}
F_2(c_2) &= c_2 + \int_{-\infty}^{\infty} \sin(c_1)p(c_1|c_2) \dif c_1\\
&= c_2 + \sin(\log(x_2))\,.
\end{split}
\label{supp.eq.F2_with_corr}
\end{equation}
\end{small}

Therefore, considering the relationship between $c_1$ and $c_2$ is crucial when using either covariate to interpret the population trend.

In summary, disentangled covariate effects of NAMs, combined with those effects contributed by covariate dependence, shape human-understandable explanations aligned with population trends. \emph{While ignoring potential dependencies in NAMs may not impact prediction performance, it can result in ambiguous or misleading interpretations when analyzing population trends.}

\subsection{Covariate Marginalization}
\label{supp.subsubsec.how_marg}
This section introduces our proposed marginalization approach to improve the trustworthiness of NAMs when trying to understand the dependency of a covariate on the response.  The dependency of covariates can be modeled with a multivariate Gaussian distribution as $p(\boldsymbol{c}|c_i)=\mathcal{N}(\hat{\boldsymbol{\mu}}(c_i), \hat{\boldsymbol{\Sigma}}(c_i))$ where $\hat{\boldsymbol{\mu}}(c_i)$ represents the mean vector and  $\hat{\boldsymbol{\Sigma}}(c_i)$ the covariance matrix conditioned on $c_i$. From $p(\boldsymbol{c}|c_i)$, one can extract the distribution of an individual covariate $c_k$ conditioned on $c_i$, as $p(c_k|c_i)=\mathcal{N}(\hat{\mu}_k(c_i), \hat{\Sigma}_{k,k}(c_i))$, e.g., how age $c_i$ determines weight $c_k$. One can also extract the joint distribution of $c_{K_1}$ and $c_{K_2}$ conditioned on $c_i$ as $p(c_{K_1}, c_{K_2}|c_i)$ from $p(\boldsymbol{c}|c_i)$, i.e.,
\begin{small}
\begin{equation}
\begin{split}
p(c_{K_1}, c_{K_2}|c_i) =
\mathcal{N}
\left(
\begin{bmatrix}
\hat{\mu}_{K_1}(c_i) \\
\hat{\mu}_{K_2}(c_i) 
\end{bmatrix},
\begin{bmatrix}
\hat{\Sigma}_{K_1, K_1}(c_i) & \hat{\Sigma}_{K_1, K_2}(c_i)  \\
\hat{\Sigma}_{K_2, K_1}(c_i) & \hat{\Sigma}_{K_2, K_2}(c_i)
\end{bmatrix}
\right),
\end{split}
\label{supp.eq.mgd}
\end{equation}
\end{small}
where $\hat{\Sigma}_{K_1, K_2}(c_i)$ is the covariance between $c_{K_1}$ and $c_{K_2}$. 

If the covariates are independent with each other, $\hat{\boldsymbol{\Sigma}}$ is a diagonal matrix. 

We employ subnetworks $\{g_i(c_i)\}$, each controlled by an individual covariate $c_i$, to model the corresponding conditional distributions $\{p(\boldsymbol{c}|c_i)\}$. Specifically, each subnetwork $g_i(c_i)$ captures a multivariate Gaussian distribution, expressed as: $p(\boldsymbol{c}|c_i) = \mathcal{N}(g^m_i(c_i), g^v_i(c_i))$, where the mean vector $g^m_i(c_i)$ and the covariance matrix $g^v_i(c_i)$ are predicted by \(g_i(c_i)\), as illustrated in Fig.~\ref{fig.method}. 

Next, from Eq.~\eqref{supp.eq.namlss}, the observation $y$ can be formulated as 
\begin{small}
\begin{equation}
y=f^m(\boldsymbol{c},x) + f^v(\boldsymbol{c},x)\cdot \epsilon \\, ~\epsilon \sim \mathcal{N}(0,1) \\.
\end{equation}
\end{small}

With trained $\{f_i\}$ and $\{g_i\}$, we now investigate how $c_i$ influences the distribution of the observation $y$ as $p(y|c_i,x)= \mathcal{N}(\Tilde{\mu}(c_i, x), \Tilde{\sigma}^2(c_i, x))$, where $\Tilde{\mu}(c_i, x)$ is the expectation of $y$ when fixing $c_i$ and $x$, i.e. $\mathrm{E}[y|c_i,x]$; and $\Tilde{\sigma}^2(c_i, x)$ is the variance of $y$ when fixing $c_i$ and $x$, i.e. $\mathrm{Var}(y|c_i,x)$.


\paragraph{Mean of $p(y|c_i,x)$.}
We expand the two variable case in Sec.~\ref{subsec.why_mrg} to multi-covariates, with the \emph{law of total expectation}~(1)
\begin{small}
\begin{equation}
\begin{split}
\Tilde{\mu}(c_i, x) &= \mathrm{E}[y|c_i,x] = \mathrm{E}[(f^m(\boldsymbol{c},x) + f^v(\boldsymbol{c},x)\cdot \boldsymbol{\epsilon})|c_i,x] \\& \overset{(1)}{=} \mathrm{E}[f^m(\boldsymbol{c},x)|c_i,x] = \int_{-\infty}^{\infty} f^m(\boldsymbol{c},x) p(\boldsymbol{c}_{k \neq i}|c_i) \dif \boldsymbol{c}_{k \neq i} \\
&= \int_{-\infty}^{\infty} (\sum_{k=1}^{N} f^m_{k}(c_{k},x)) p(\boldsymbol{c}_{k \neq i}|c_i) \dif \boldsymbol{c}_{k \neq i} \\
&= f^m_i(c_i,x) + \underbrace{\int_{-\infty}^{\infty} (\sum_{k \neq i} f^m_{k}(c_{k},x)) p(\boldsymbol{c}_{k \neq i}|c_i) \dif \boldsymbol{c}_{k \neq i}}_{:=H} 
\label{supp.eq.E_y_given_ci_1}
\end{split}
\end{equation}
\end{small}

where $f^m_k(c_k)$ represents the interpretation from the additive subnetwork $f_i$ of \texttt{LucidAtlas}, while $H$ accounts for the contributions from the dependencies between the covariates which can be further simplified as follows  
\begin{small}
\begin{equation}
\begin{split}
& H = \sum_{k \neq i} \int_{-\infty}^{\infty} f^m_{k}(c_{k},x) p(\boldsymbol{c}_{k \neq i}|c_i) \dif \boldsymbol{c}_{k \neq i} \\
&= \sum_{k \neq i} \int_{-\infty}^{\infty} f^m_{k}(c_{k},x)(\int_{-\infty}^{\infty}  p(\boldsymbol{c}_{j \neq \{i,k\}}, c_k|c_i) \dif \boldsymbol{c}_{j \neq \{i,k\}}) \dif c_k \\
&= \sum_{k \neq i} \int_{-\infty}^{\infty} f^m_{k}(c_{k},x) p(c_k|c_i) \dif c_k 
\label{supp.eq.H}
\end{split}
\end{equation}
\end{small}
where $\boldsymbol{c}_{k \neq i}=[c_1, ..., c_{i-1}, c_{i+1}, ..., c_N]$.

Eq.~\eqref{supp.eq.H} indicates that even when multiple covariates are involved, only conditional dependencies with respect to individual covariates ($p(c_k|c_i)$) are required to compute $\Tilde{\mu}(c_i, x)$ as a consequence of the additive model for a NAM, which simplifies computations.

Therefore, 
\begin{small}
\begin{equation}
\begin{split}
& \Tilde{\mu}(c_i, x) = f^m_i(c_i,x) + \sum_{k \neq i} \int_{-\infty}^{\infty} f^m_{k}(c_{k},x) p(c_k|c_i) \dif c_k\,. 
\label{supp.eq.E_y_given_ci}
\end{split}
\end{equation}
\end{small}

\paragraph{Variance of $p(y|c_i,x)$.}

The \emph{law of total variance} is $\mathrm{Var}(Y) =\mathrm{E}[\mathrm{Var}(Y|X)] + \mathrm{Var}(\mathrm{E}[Y|X])$ which states that the total variance of a random variable $Y$ can be broken into two parts: \textcircled{1} the \textbf{expected variance of $Y$ given $X$}, which represents how much $Y$ fluctuates around its mean for each specific value of $X$; and \textcircled{2} The variance of the \textbf{expected value of $Y$ given $X$}, which measures how much the mean of $Y$ changes as $X$ varies.
With the \emph{law of total variance}, 
\begin{small}
\begin{equation}
\mathrm{Var}(y|c_i,x)= \underbrace{\mathrm{E}[\mathrm{Var}(y|\boldsymbol{c}_{k \neq i}, c_i,x)]}_{ \text{\textcircled{1}} :=\Tilde{\sigma}^2_E(c_i, x)} + \underbrace{\mathrm{Var}(\mathrm{E}[y|\boldsymbol{c}_{k \neq i}, c_i,x])}_{\text{\textcircled{2}}:=\Tilde{\sigma}^2_V(c_i, x)}
\label{supp.eq.Var_dist}
\end{equation}
\end{small}
The expected variance of $f^v(\boldsymbol{c}, x)$ given $c_i$ and $x$ is
\begin{small}
\begin{equation}
\begin{split}
&\Tilde{\sigma}^2_E(c_i, x) = \mathrm{E}[\mathrm{Var}(y|\boldsymbol{c}_{k \neq i}, c_i,x)] = \mathrm{E}[f^v(\boldsymbol{c},x)|c_i,x]  \\
&= \int_{-\infty}^{\infty} f^v(\boldsymbol{c},x) p(\boldsymbol{c}_{k \neq i}|c_i) d\boldsymbol{c}_{k \neq i} \\
&= f^v_i(c_i,x) + \sum_{k \neq i} \int_{-\infty}^{\infty} f^v_{k}(c_{k},x) p(c_k|c_i) \dif c_k\,.
\end{split}
\label{supp.eq.E_of_V}
\end{equation}
\end{small}

And the variance of the expected value of $f^m(\boldsymbol{c},x)$ given $c_i$ and $x$ can be computed as
\begin{small}
\begin{equation}
\begin{split}
&\Tilde{\sigma}^2_V(c_i, x) =\mathrm{Var}(\mathrm{E}[y|\boldsymbol{c}_{k \neq i}, c_i,x]) = \mathrm{Var}(f^m(c_i, \boldsymbol{c}_{k \neq i},x)|c_i) \\
&=\mathrm{Var}(f^m_i(c_i, x) + \sum_{k \neq i}f^m_k(c_k, x)|c_i, x) \\
&=\mathrm{Var}(\sum_{k \neq i}f^m_k(c_k, x)|c_i, x) \\ &= \sum_{k \neq i} \underbrace{\mathrm{Var}(f^m_k(c_k,x)|c_i, x)}_{\text{\textcircled{3}}} \\ &+ \mathop{\sum\sum}_{K_1\neq K_2 \neq i} \underbrace{\mathrm{Cov}(f^m_{K_1}(c_{K_1}, x), f^m_{K_2}(c_{K_2}, x) | c_i, x)}_{\text{\textcircled{4}}}
\end{split}
\label{supp.eq.V_of_E}
\end{equation}
\end{small}
where
\begin{small}
\begin{equation}
\begin{split}
&\text{\textcircled{3}} = \int_{-\infty}^{\infty} (f^m_k(c_k, x) - \Tilde{\mu}_k(c_i, x))^2p(c_k| c_i) \dif c_k ,\\
& \Tilde{\mu}_k(c_i, x) = \int_{-\infty}^{\infty} f^m_k(c_k, x)p(c_k|c_i)dc_k
\end{split}
\label{supp.eq.V_of_E_part1}
\end{equation}
\end{small}

\begin{equation}
\begin{split}
\text{\textcircled{4}} = \int_{-\infty}^{\infty}\int_{-\infty}^{\infty} f^m_{K_1}(c_{K_1}, x)f^m_{K_2}(c_{K_2}, x)p(c_{K_1}, c_{K_2}|c_i)\dif c_{K_1} \dif c_{K_2} \\
 - \Tilde{\mu}_{K_1}(c_i, x) \Tilde{\mu}_{K_2}(c_i, x) 
\end{split}
\label{supp.eq.V_of_E_part2}
\end{equation}


Eqs.~\eqref{supp.eq.E_of_V}-\eqref{supp.eq.V_of_E_part2} imply that instead of sampling the entire covariate space, one only needs to sample from the joint Gaussian distribution between the two covariates, conditioned on the individual covariates, to obtain the marginalized distribution $p(y|c_i,x)$.

\paragraph{Approximation.} 
The integrals in $\Tilde{\mu}(c_i, x)$ (in Eq.~\eqref{supp.eq.E_y_given_ci}), $\Tilde{\sigma}^2_E(c_i, x)$ (in Eq.~\eqref{supp.eq.E_of_V}) and $\Tilde{\sigma}^2_V(c_i, x)$ (in Eqs.~\eqref{supp.eq.E_of_V}-\eqref{supp.eq.V_of_E_part2}) can be approximated using Monte Carlo sampling. E.g. for $\Tilde{\mu}(c_i, x)$, for each covariate $c_k$ one can sample $L$ values $\{\hat{c}_k^l\}_{l=1, ..., L}$ from the distribution of covariates $p(c_k|c_i)$ given by $g_k(\cdot)$ to approximate $\Tilde{\mu}(c_i, x)$ as  
\begin{small}
\begin{equation}
\begin{split}
\Tilde{\mu}(c_i, x) \approx f^m_i(c_i) + \frac{1}{L} \sum_{k \neq i}\sum_{l=1}^{L} f^m_k(\hat{c}_{k}^{l},x)\,,
\label{supp.eq.sample_for_num}
\end{split}
\end{equation}
\end{small}  
where the $\{\hat{c}_{k}^{l}\}_{l=1}^{L}$ are sampled from $p(c_k|c_i)$.

\paragraph{Computational Complexity.}
Suppose there are $N$ covariates and $L$ samples. The computational complexity of marginalizing the NAM for a covariate is \(\mathcal{O}(LN)\), making it feasible in practice. In contrast, for a black-box model, which does not assume our additive structure, the complexity is exponentially higher at \(\mathcal{O}(L^N)\), making direct computation infeasible for large \(N\).

As a result, we obtain $\Tilde{\mu}(c_i, x)$ and $\Tilde{\sigma}^2(c_i, x)=\Tilde{\sigma}^2_E(c_i, x)+\Tilde{\sigma}^2_V(c_i, x)$ to parameterize $p(y|c_i) = \mathcal{N}(\Tilde{\mu}(c_i, x), \Tilde{\sigma}^2(c_i, x))$, capturing the influence of a single covariate $c_i$ on the observation $y$. Our approach aligns with NAM interpretations and can be applied post-hoc.

\subsubsection{Imputation}
\label{supp.sec.imputation}
Our approach naturally facilitates the imputation of missing covariates, as it inherently predicts the conditional distributions \(\{p(c_k \mid c_i)\}\), enabling a principled way to estimate missing values.
 Specifically, if $c_i$ is missing, one can choose the $g_s$ whose uncertainty is the smallest as the predictor for $c_i$ as $s \gets \arg\min_{k,k \neq i} \{{g^v_{k,i}(c_k)}\}$.

\subsubsection{Individualized Prediction}
\label{supp.sec.ind_pred}
One challenge in the context of atlas discovery is to make individualized predictions when observations are predominantly limited to a single time point, i.e., when the atlas is built from cross sectional data. \texttt{LucidAtlas} provides an approach for individualized prediction based on previous observations. Note that this approach is not based on true longitudinal data  (as such data is frequently not available) but instead aims to predict individual future responses based on the cross-sectional population trend.

We define our problem as follows. Given an observation $y^t$ at $x$ with its corresponding covariates $\boldsymbol{c}^t$  at time $t$, how will a subject's response change when $\boldsymbol{c}^t$ changes to $\boldsymbol{c}^{t+1}$ at time $t+1$?

First, we can obtain the probability distribution with  \texttt{LucidAtlas}, as $p(y^t|\boldsymbol{c}^t,x)=\frac{1}{\sqrt{2\pi f^v(\boldsymbol{c}^t,x)}}\exp(-\frac{(y^t-f^m(\boldsymbol{c}^t,x))^2}{2 \cdot f^v(\boldsymbol{c}^t,x)})$.

\begin{assumption}
We assume that the percentile of a subject remains stationary between  observations at two nearby time points, i.e., the cumulative distribution, $\mathrm{F}$, should be stationary: $\mathrm{F}(y^t) = \mathrm{F}(y^{t+1})$.
\label{supp.asp.sig}
\end{assumption}
An intuitive example for the Assumption~\ref{supp.asp.sig} is that if a child has the largest airway among all 2-year-olds, it is likely that this child's airway will remain the largest over a short period of time. Now we have
\begin{small}
\begin{equation}
\begin{split}
\mathrm{F}(y^t)=\frac{1}{2}[1+\mathrm{erf}(\frac{y^t-f^m(\boldsymbol{c}^t,x)}{\sqrt{2f^v(\boldsymbol{c}^t,x)}})]\,, \\
\mathrm{F}(y^{t+1})=\frac{1}{2}[1+\mathrm{erf}(\frac{y^{t+1}-f^m(\boldsymbol{c}^{t+1},x)}{\sqrt{2f^v(\boldsymbol{c}^{t+1},x)}})]\,.
\end{split}
\label{supp.eq.F_cum}
\end{equation}
\end{small}
\begin{small}
\begin{equation}
\begin{split}
\mathrm{F}(y^t) = \mathrm{F}(y^{t+1}) \Rightarrow  \frac{y^t-f^m(\boldsymbol{c}^t,x)}{\sqrt{2f^v(\boldsymbol{c}^t,x)}} = \frac{y^{t+1}-f^m(\boldsymbol{c}^{t+1},x)}{\sqrt{2f^v(\boldsymbol{c}^{t+1},x)}}  \Rightarrow \\
y^{t+1} = f^m(\boldsymbol{c}^{t+1},x) + {\sqrt{\frac{f^v(\boldsymbol{c}^{t+1},x)}{f^v(\boldsymbol{c}^{t},x)}}(y^t-f^m(\boldsymbol{c}^t,x))}\,.
\end{split}
\end{equation}
\end{small}

Therefore, an approximate individualized prediction can be obtained as $y^{t+1} \approx f^m(\boldsymbol{c}^{t+1},x) + {\sqrt{\frac{f^v(\boldsymbol{c}^{t+1},x)}{f^v(\boldsymbol{c}^{t},x)}}(y^t-f^m(\boldsymbol{c}^t,x))}$\,.

\section{Network Architecture}

\begin{figure*}
    \centering
    \includegraphics[width=0.6\linewidth]{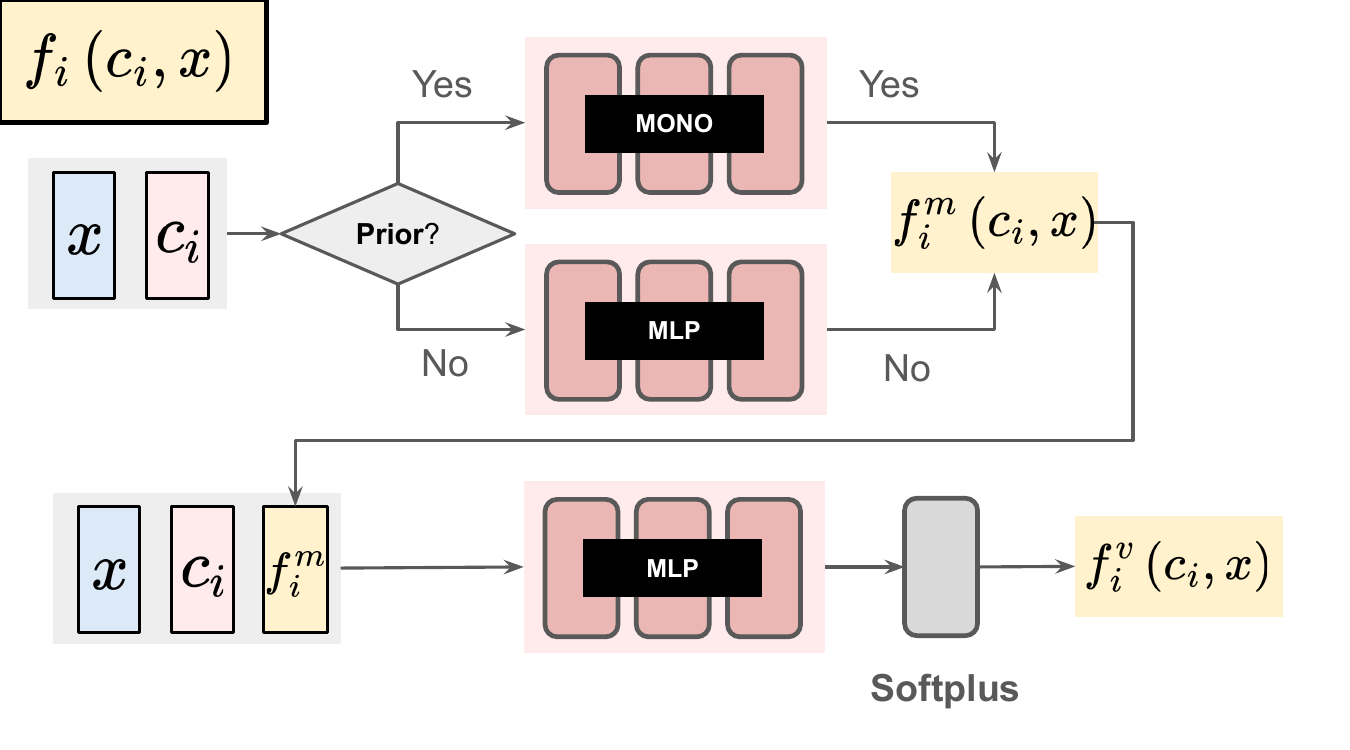}
    \caption{Network Architecture of Additive Subnetwork $f_i$ in \texttt{LucidAtlas}. }
    \label{supp.fig.f_i}
\end{figure*}

\begin{figure*}
    \centering
    \includegraphics[width=0.6\linewidth]{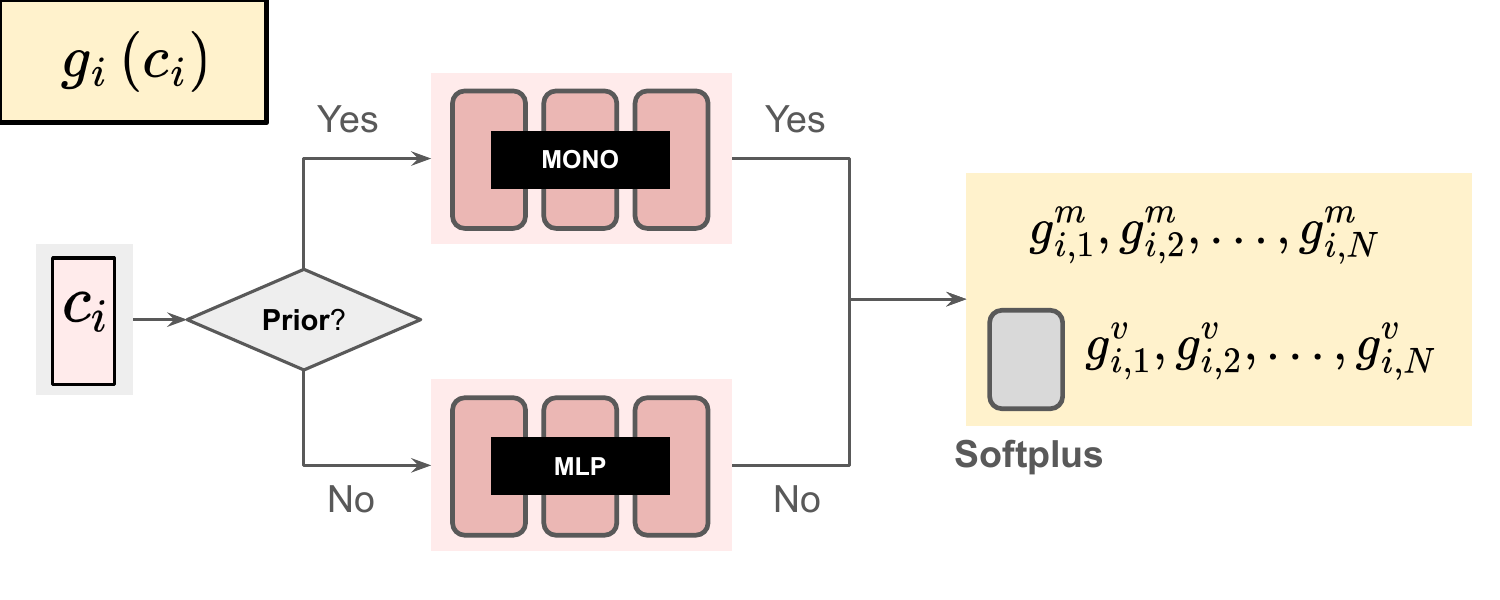}
    \caption{Network Architecture of $g_i$ in \texttt{LucidAtlas} for Modeling Covariate Dependencies.}
    \label{supp.fig.g_i}
\end{figure*}

Fig.~\ref{supp.fig.f_i} shows the network architecture of the additive subnetwork $f_i$, which receives the anatomical location $x$ and covariate $c_i$ to predict the additive contribution $f^m_i(c_i, x)$ and $f^v_i(c_i, x)$ to the mean and variance for the distributional parameters $\mu(\boldsymbol{c}, x)$ and $\sigma^2(\boldsymbol{c}, x)$ respectively. Specifically, if there is prior knowledge, we use a monotonic neural network~\citep{kitouni2023mono} as the backbone to predict $f^m_i(c_i, x)$; if there is no prior knowledge, we use an MLP to predict $f^m_i(c_i, x)$. Another MLP receives the $f^m_i(c_i, x)$ with $c_i$ and $x$ to predict the contribution $f^v_i(c_i, x)$ to the 
 overall variance $\sigma^2(\boldsymbol{c}, x)$. Considering that the variance should be a number $\geq 0$, a $softplus$ activation layer is used at the output of the MLP for $f_v^i(c_i, x)$ to ensure $f_v^i(c_i, x)$ is a non-negative number.

Fig.~\ref{supp.fig.g_i} shows the network architecture for the dependence modeling network $g_i$, which receives the covariate $c_i$ to predict the parameters of the conditional distributions $\{p(c_k| c_i)\}$, whose distributional parameters are $\{g^m_{i,k}\}$ for means and $\{g^v_{i,k}\}$ for variances. One can also use monotonic neural networks if prior knowledge exists about the covariate dependence. For example, for children, on average, weight increases with age. Suppose there is no prior knowledge about monotonicity. In that case, one can use $\operatorname{Lip}^1$ constrained network discussed in~\cite{kitouni2023mono} for modeling covariates dependence, which prevents sharp and abrupt changes which are not likely to happen.

\section{Datasets}
\label{supp.sec.dataset}

\subsection{OASIS Brain}
\label{supp.subsec.brain_dataset}
The Open Access Series of Imaging Studies (OASIS) is a project aimed at making MRI data sets of the brain freely available to the scientific community~\citep{marcus2007oasis}. 

The OASIS Brain dataset we use is publicly available in a preprocessed form~\footnote{\url{https://www.kaggle.com/datasets/jboysen/mri-and-alzheimers}}. The OASIS Brain dataset consists of two sets, i.e., 

\begin{itemize}
    \item[1] \textbf{A Cross-Sectional MRI Dataset (416 Subjects, Ages 18–96).} 
    100 of the included subjects are over the age of 60 and have been clinically diagnosed with very mild to moderate Alzheimer’s disease (AD). Additionally, a reliability data set is included containing 20 nondemented subjects imaged on a subsequent visit within 90 days of their initial session.
    \item[2] \textbf{A Longitudinal MRI Dataset in Nondemented and Demented Older Adults (150 Subjects, Ages 60–96).}
    This set consists of a longitudinal collection of 150 subjects aged 60 to 96. Each subject was scanned on two or more visits, separated by at least one year for a total of 373 imaging sessions. For each subject, 3 or 4 individual T1-weighted MRI scans obtained in single scan sessions are included. The subjects are all right-handed and include both men and women. 72 of the subjects were characterized as nondemented throughout the study. 64 of the included subjects were characterized as demented at the time of their initial visits and remained so for subsequent scans, including 51 individuals with mild to moderate Alzheimer’s disease. Another 14 subjects were characterized as nondemented at the time of their initial visit and were subsequently characterized as demented at a later visit.
\end{itemize}

Our experiments include four covariates: age, socioeconomic status (SES), mini-mental state examination (MMSE), and clinical dementia rating (CDR). The outcome variable is normalized whole brain volume (nWBV), which is a scalar.  

\emph{We aim to investigate the relationships between these covariates and brain volume.} Based on prior knowledge, the atlas brain volume should not increase with age or CDR, nor decrease when mental state improves.

\subsection{Pediatric Airway}
\label{supp.subsec.airway_dataset}

\begin{table}[htbp!]
\centering
\begin{tabular}{l r r r r r r r r r r} 
\toprule
    \# observations     &  1 & 2 & 3 & 4 & 5 & 6 & 7 & 9 & 11\\ 
\hline\hline
    \# patients   &  230 & 12 &  6 & 8 & 3 & 2 & 1 & 1 & 1 \\
\bottomrule
\end{tabular}
\caption{Number of patients for a given number of observations for the pediatric airway dataset. For example, the 1st column indicates that there are 230 patients who were only observed once. }
\label{tab.num_of_observations}
\end{table}

\begin{table*}

\resizebox{\textwidth}{!}{%
\begin{tabular}{p{0.11\textwidth}p{0.11\textwidth}p{0.11\textwidth}p{0.11\textwidth}p{0.11\textwidth}p{0.11\textwidth}p{0.11\textwidth}p{0.11\textwidth}p{0.11\textwidth}p{0.11\textwidth}p{0.11\textwidth}p{0.11\textwidth}}
\toprule
P- &      0&      10  &      20  &      30  &      40  & 50 & 60 & 70 & 80 & 90 & 100\\
\midrule
 &
\includegraphics[width=0.2\columnwidth]{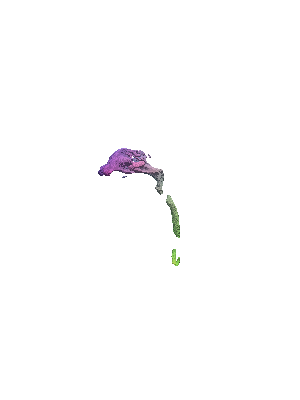} &  
\includegraphics[width=0.2\columnwidth]{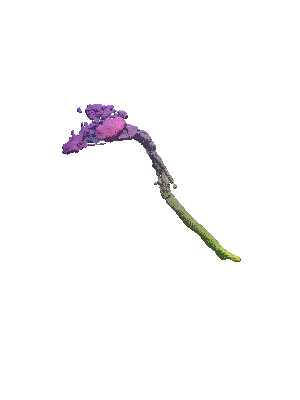} &  
\includegraphics[width=0.2\columnwidth]{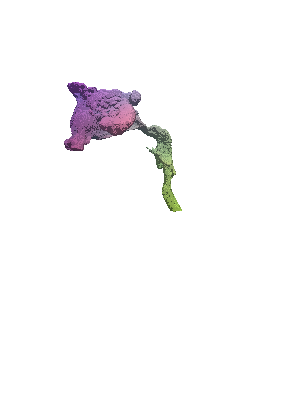} &  
\includegraphics[width=0.2\columnwidth]{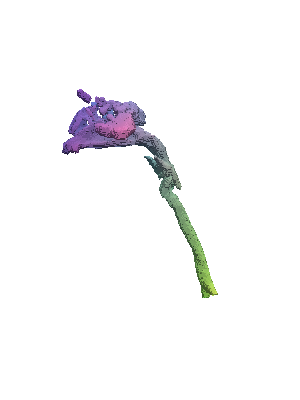} &  
\includegraphics[width=0.2\columnwidth]{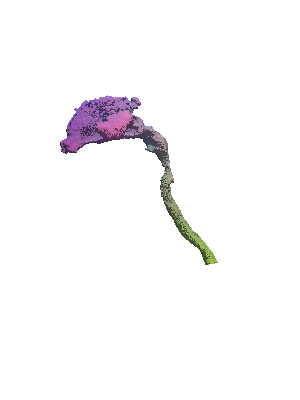} & 
\includegraphics[width=0.2\columnwidth]{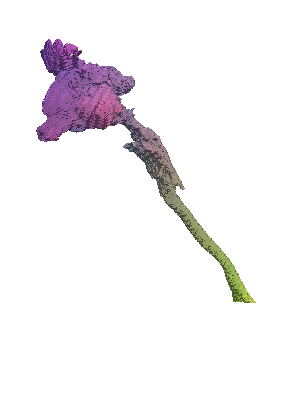} &  
\includegraphics[width=0.2\columnwidth]{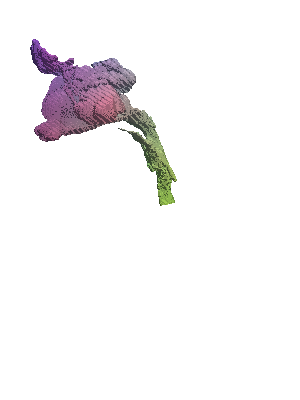} &  
\includegraphics[width=0.2\columnwidth]{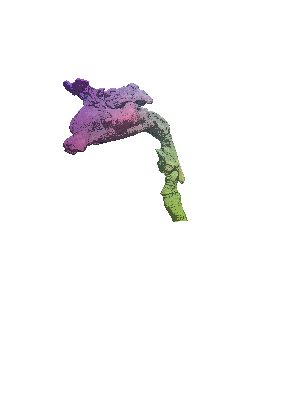} &  
\includegraphics[width=0.2\columnwidth]{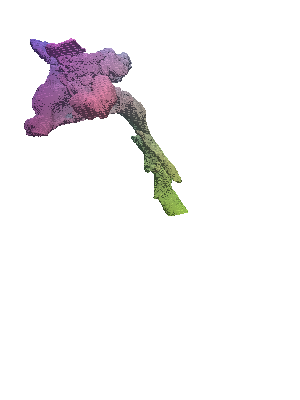} &  
\includegraphics[width=0.2\columnwidth]{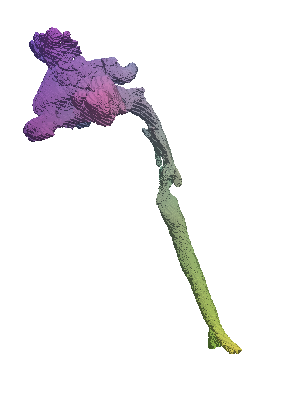} &
\includegraphics[width=0.2\columnwidth]{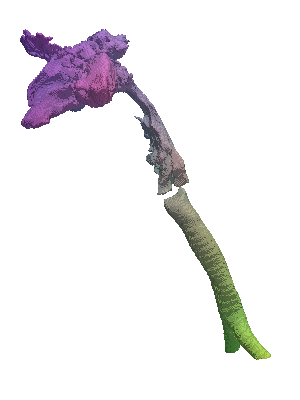} \\

\bottomrule
\end{tabular}}
\resizebox{\textwidth}{!}{%
\begin{tabular}{p{0.06\textwidth}p{0.06\textwidth}p{0.06\textwidth}p{0.06\textwidth}p{0.06\textwidth}p{0.06\textwidth}p{0.06\textwidth}p{0.06\textwidth}p{0.06\textwidth}p{0.06\textwidth}p{0.06\textwidth}p{0.06\textwidth}}
\toprule
P- &      0&      10  &      20  &      30  &      40  & 50 & 60 & 70 & 80 & 90 & 100\\
\midrule

age         &     1.00 &    23.00 &    55.00 &    71.00 &    89.00 &   111.00 &   129.00 &   161.00 &   179.00 &   199.00 &   233.00 \\
m-vol &     4.56 &    16.84 &    29.53 &    28.91 &    27.31 &    70.90 &    71.23 &    43.34 &    78.63 &   102.35 &   113.84 \\

\bottomrule

\end{tabular}}
\caption{Visualization and demographic information of our 3D airway shape dataset. Shapes of $\{0, 10, 20, 30, 40, 50, 60, 70, 80, 90, 100\}$-th age percentiles 
 are plotted with their covariates (age/month) printed in the table. M-vol (measured volume) is the volume ($cm^3$) of the gold standard shapes based on the actual imaging. }
\label{supp.tab.vis_demo_dataset}
\end{table*}

The airway shapes are extracted from computed tomography (CT) images. The real CT images are from children ranging in age from 1 month to $\sim$19 years old. Acquiring CT images is costly. Further, CT uses ionizing radiation which should be avoided, especially in children, due to cancer risks. Hence, it is difficult to acquire such CTs for many children. Instead, the data was acquired by serendipity from children who received CTs for reasons other than airway obstructions (e.g., because they had cancer)~\citep{jiao2023naisr}. This also explains why it is difficult to acquire longitudinal data. E.g., one of the patients has 11 timepoints because a very sick child was scanned 11 times.

The pedatric airway dataset includes 230 cross-sectional observations (where a patient was only imaged once) and 34 longitudinal observations. 176 patients (i.e., 263 shapes) have all three covariates (age, weight, height) and 11 annotated anatomical landmarks. 6 landmarks are located on the upper airway section for this experiment. Errors in the shapes $\{\mathcal{S}^k\}$ may arise from image segmentation error, differences in head positioning, missing parts of the airway shapes due to incomplete image coverage, and dynamic airway deformations due to breathing. Table~\ref{tab.num_of_observations} shows the distribution of the number of observations across patients. Most of the patients in the dataset only have one observation; only 22 patients have $\geq 3$ observation times. Table~\ref{supp.tab.vis_demo_dataset} shows the shapes and demographic information at different age percentiles for the whole data set. Similar to the OASIS Brain dataset, the time span of the longitudinal data for each patient is far shorter than the time span across the entire dataset.

\subsubsection{Data Preparation for Pediatric Airway Atlas}

The image processing pipeline includes three steps: 1) automatic airway segmentation from CT images; 2) airway representation with a centerline and cross sections.

\paragraph{Airway Segmentation.}
A deep learning-based approach is used for automatic upper airway segmentation from CT images. The segmentation model is trained in two steps. The first step predicts the segmentation using a coarse version of the scans. The second step makes the segmentation prediction on original images. This step takes in the image as input, but also uses the first step prediction as an additional input. Each step is implemented as a U-Net~\citep{unetcciccek20163d, unetronneberger2015u}.

The automatic segmentation model is developed based on a dataset containing 68 pairs of airway CT images and their corresponding manual segmentations.

\paragraph{Centerline and Cross Sections.}

The pediatric airway dataset is constructed by extracting 358 airway geometries from CT images with our automatic segmentation approach. The upper airways, like any tube-like structures, can be approximated by a centerline with cross sections~\citep{atlashong2013pediatric}. Following the approach in~\citep{atlashong2013pediatric}, the airway centerline is inferred based on the heat distribution along the airway provided by solving Laplace's equation. The iso-surfaces of heat values are extracted from the Laplace solution and the centerlines are considered as the centers of the iso-surfaces. Cross sections are cut from the airway geometry using planes that are orthogonal to the tangent of the centerline.

\paragraph{Pediatric Airway Atlas Construction.}
Similar as the approach in~\citep{atlashong2013pediatric}, the cross-sectional area is considered as the airway's main feature. For each point on the centerline, it has a distance $x$ from the nasal spine which is normalized to 1 over the length of the airway, and a cross-sectional area $y$. The 1D function for airway geometry is the curve $c(x)$ that smoothly passes through all these points on the centerline, as $y=c(x)$.

The airway curves are aligned based on six key anatomic landmarks $\{\boldsymbol{p}_i\}$: nasal spine, choana, epiglottic tip, true vocal cord (TVC), subglottis, and carina.


Each landmark $\boldsymbol{p}_{i}=\left(p_{ix}, p_{iy}, p_{iz}\right)$ is projected onto the centerline to obtain the corresponding depth $x_{i}$ along the centerline. For example, the depth of nasal spine $x_{nasal spine}$ should be at 0 while the depth of carina $x_{\text {carina}}$ should be at 1. For each landmark, there is a mean position $\overline{\boldsymbol{p}}_{i}=\left(\bar{p}_{ix}, \bar{p}_{iy}, \bar{p}_{iz}\right)$ and the mean depth $\bar{d}_{i}$ of that landmark over all cases.

A landmark-based curve registration approach ~\citep{atlashong2013pediatric} is used to estimate a piece-wise linear warping function $h_{k}(\cdot)$ for each curve $c_{k}(\cdot)$, which is strictly monotonic and places the landmark points for a particular subject $k$ at the mean location of these landmarks in the atlas, $x_{i}=h_{k}\left(\bar{x}_{i}\right)$. With the constructed warping functions, curves can then be resampled to the normalized coordinate system with $C_{k}(x)=c_{k}\left(h_{k}(x)\right)$.

\section{Experimental Setup}

\begin{table*}[]
\centering
\resizebox{0.99\textwidth}{!}{%
\begin{tabular}{|l|cccccc|}
\hline
\multirow{3}{*}{Model} & \multicolumn{4}{c|}{\texttt{LucidAtlas}}                             & \multicolumn{1}{c|}{NAMLSS} & \multirow{3}{*}{MLP+NLL} \\ \cline{2-6}
                       & \multicolumn{2}{c|}{$f^m_i$ or $f^v_i$} & \multicolumn{2}{c|}{$g_i$} & \multicolumn{1}{c|}{$f_i$}  &                          \\ \cline{2-6}
 &
  \multicolumn{1}{c|}{Monotonic} &
  \multicolumn{1}{c|}{MLP} &
  \multicolumn{1}{c|}{Monotonic} &
  \multicolumn{1}{c|}{$Lip^1$ Constrained} &
  \multicolumn{1}{c|}{MLP} &
   \\ \hline
Layers &
  \multicolumn{1}{c|}{$[D_{in}, 128, D_{out}]$} &
  \multicolumn{1}{c|}{$[D_{in}, 128, D_{out}]$} &
  \multicolumn{2}{c|}{$[D_{in}, 128, D_{out}]$} &
  \multicolumn{1}{c|}{$[D_{in}, 128, D_{out}]$} &
  $[D_{in}, 128, D_{out}]$ \\ \hline
Activation &
  \multicolumn{1}{c|}{GroupSort} &
  \multicolumn{1}{c|}{GeLU} &
  \multicolumn{2}{c|}{GroupSort} &
  \multicolumn{1}{c|}{GroupSort} &
  GeLU \\ \hline
Learning Rate          & \multicolumn{6}{c|}{1e-2}                                                                                                     \\ \hline
Num of Epoch           & \multicolumn{6}{c|}{500}                                                                                                      \\ \hline
Output Activation      & \multicolumn{6}{c|}{Linear, Softplus}                                                                                         \\ \hline
Others                 & \multicolumn{6}{c|}{Adam optimizer, CosineAnnealingLR, Earlystopping}                                                         \\ \hline
\end{tabular}
}
\caption{Hyperparameter Settings for Comparison Methods. GroupSort and $Lip^1$-constrained networks were introduced in~\citep{kitouni2023mono}. Here, $D_{in}$ represents the input dimension, while $D_{out}$ denotes the output dimension. For details on $D_{in}$ and $D_{out}$, refer to Fig.~\ref{supp.fig.f_i} and Fig.~\ref{supp.fig.g_i}.}
\label{supp.tab.hyperparameter}
\end{table*}

Tab.~\ref{supp.tab.hyperparameter} illustrates the hyperparameter settings of our approach, for NAMLSS~\citep{thielmann2024namlss} and the \textit{MLP+NLL} comparison. For our approach and all comparison methods, we ues $15\%$ of the training data set by the patient as a validation set for early stopping. The batch size for the Pediatric Airway Dataset is set to $1024$, while for the OASIS Brain dataset it is set to 32. For other comparison methods, we use their publicly available implementation, which we describe in the  following.

\paragraph{NAM.} We use the official PyTorch implementation of NAM~\footnote{\url{https://github.com/lemeln/nam/tree/main?tab=readme-ov-file}}.
We evaluate the NAM using feature networks with one hidden layers (to keep consisent with hyperparameter settings in Table.~\ref{supp.tab.hyperparameter}), each containing $128$ units and employing ExU activation. A dropout rate of $0.1$ is applied, and the ensemble consists of $20$ learners. We use $15\%$ of the training data set by patient as a validation set for early stopping. All other experimental settings follow the recommended or default configurations.

\paragraph{Explainable Boosting Machine.} The explainable boosting machine (EBM) is an open-source Python implementation~\footnote{\url{https://github.com/interpretml/interpret/tree/3e810552f7fcae641bf6bd945f10c66bf56c424b}} of the gradient-boosting GAM that is available as a part of the InterpretML library~\citep{lou2013EBM, nori2019interpretmlunifiedframeworkmachine}.  We use $15\%$ of the training data set by patient as a validation set for early stopping. We use the default hyperparameter setting, because we did not find a significant improvement when tuning the hyperparameters. EBM allows for control of the monotonicity of features by using isotonic regression. We find introducing prior knowledge improves EBM's performance, and thus, we use the same prior knowledge for EBM as we used in our \texttt{LucidAtlas}.

\paragraph{LightGBM.} LightGBM is a gradient boosting framework that uses tree-based learning algorithms~\citep{ke2017lightgbm}. We use the open-source implementation~\footnote{\url{https://lightgbm.readthedocs.io/en/latest/index.html}}. We use $15\%$ of the training data set by patient as a validation set for early stopping. We find that the recommended or default configurations work well.

\paragraph{Hardware.} The deep learning models are trained on a single NVIDIA GeForce RTX 3090 GPU and an Intel(R) Xeon(R) Gold 6226R CPU.

\section{More Results and Visualizations}
\label{supp.more_vis_exp}
Table~\ref{supp.exp.single_quant_ece} presents the quantitative evaluation of population distribution estimation based on the Expected Calibration Error (ECE), demonstrating that incorporating prior knowledge enhances performance on the pediatric airway dataset.

Table~\ref{supp.exp.oasis_cov_corr} evaluates the impact of accounting for covariate dependencies in marginalization. The results highlight that considering these dependencies is crucial for accurately interpreting the effects of individual covariates in neural additive models.

\begin{figure}

    \centering
    \includegraphics[width=0.7\linewidth]{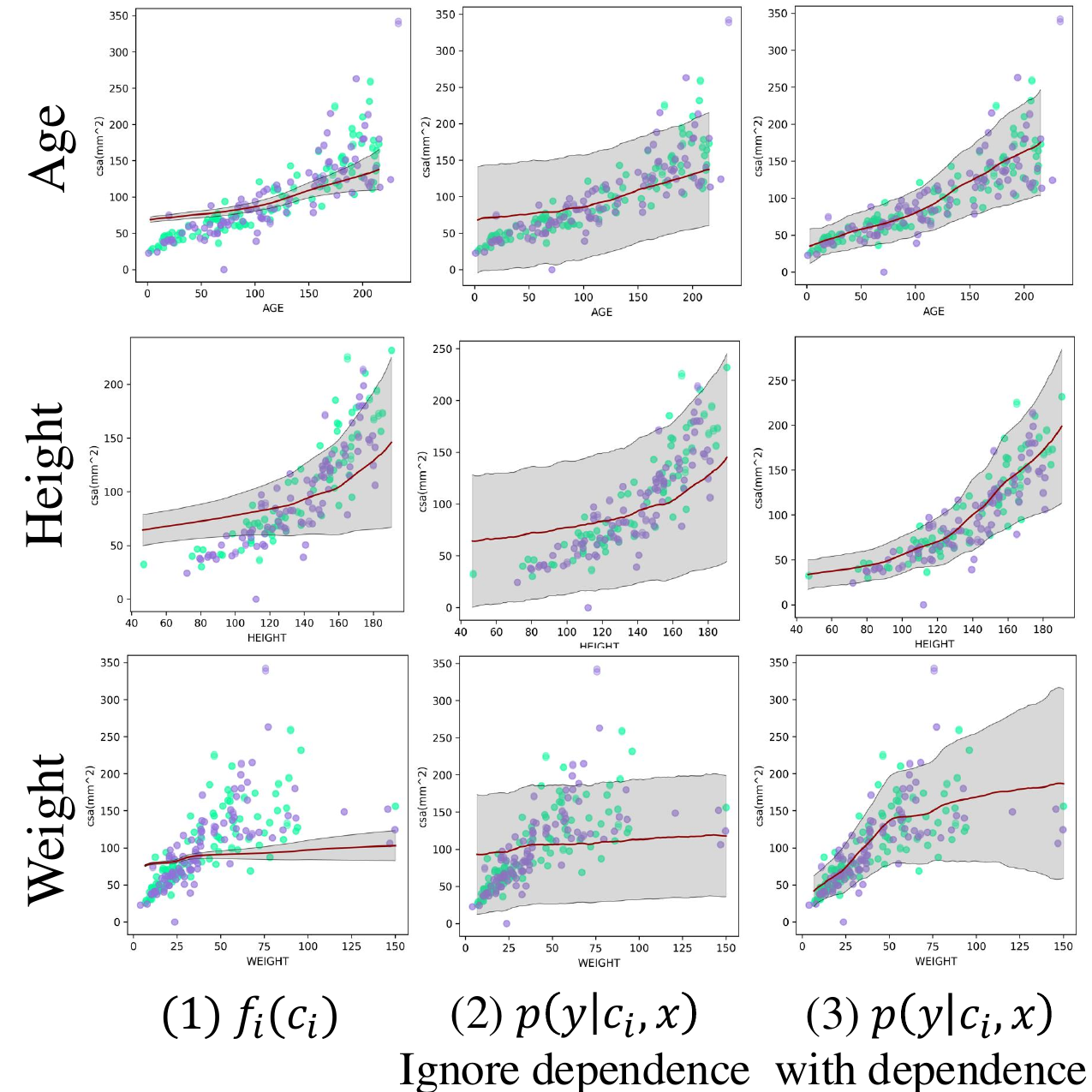}
    \caption{\small Visualizations of Covariate Interpretations from \texttt{LucidAtlas} for CSA Distribution at the Subglottis Landmark (Pediatric Airway Dataset). (1) $f_i(c_i)$ represents the disentangled covariate effect directly from a NAM as illustrated in Sec.~\ref{sec.dist_cov_effects}; (2) Marginalized covariate interpretation without accounting for covariate dependence; (3) Marginalized covariate interpretation incorporating covariate dependence. \textcolor{SeaGreen}{Green} dots denote training samples, while \textcolor{Orchid}{purple} dots indicate testing samples. The \textcolor{BrickRed}{red} lines represent the learned population trend, and the \textcolor{Gray}{gray} shading spans \(\pm 2 \times\) standard deviations. Considering covariate dependence is essential for accurately capturing how each covariate influences the population trend and associated uncertainties.} 
    \label{supp.fig.vis_whether_do_correlation_sgs}
\end{figure}

\begin{figure}
    \centering
    \includegraphics[width=0.7\linewidth]{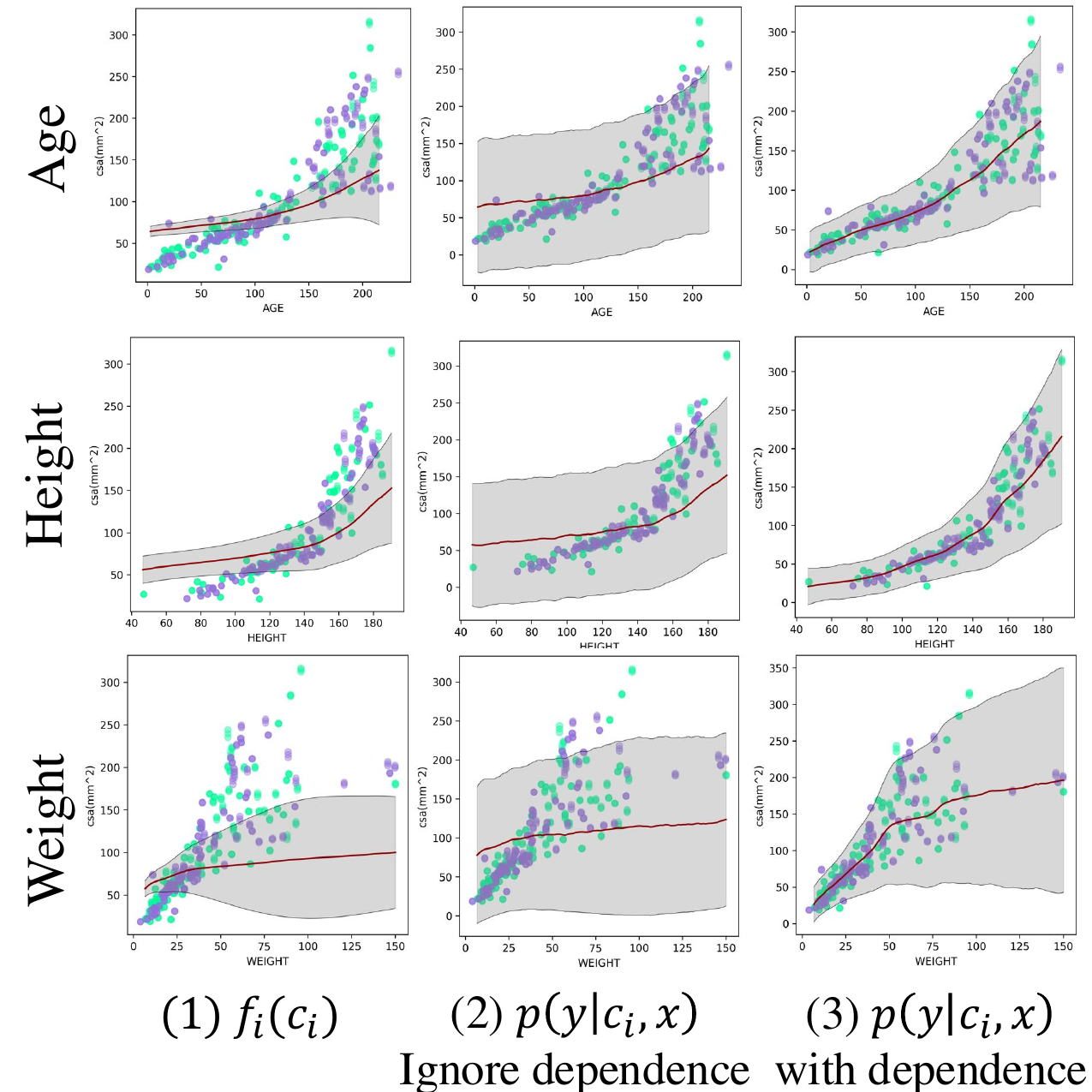}
    \caption{\small Visualizations of Covariate Interpretations from \texttt{LucidAtlas} for CSA Distribution at the Epiglottic Tip Landmark (Pediatric Airway Dataset). (1) $f_i(c_i)$ represents the disentangled covariate effect directly from a NAM as illustrated in Sec.~\ref{sec.dist_cov_effects}; (2) Marginalized covariate interpretation without accounting for covariate dependence; (3) Marginalized covariate interpretation incorporating covariate dependence. \textcolor{SeaGreen}{Green} dots denote training samples, while \textcolor{Orchid}{purple} dots indicate testing samples. The \textcolor{BrickRed}{red} lines represent the learned population trend, and the \textcolor{Gray}{gray} shading spans \(\pm 2 \times\) standard deviations. Considering covariate dependence is essential for accurately capturing how each covariate influences the population trend and associated uncertainties.} 
    \label{supp.fig.vis_whether_do_correlation_epg}
\end{figure}

\begin{figure}
    \centering
    \includegraphics[width=0.7\linewidth]{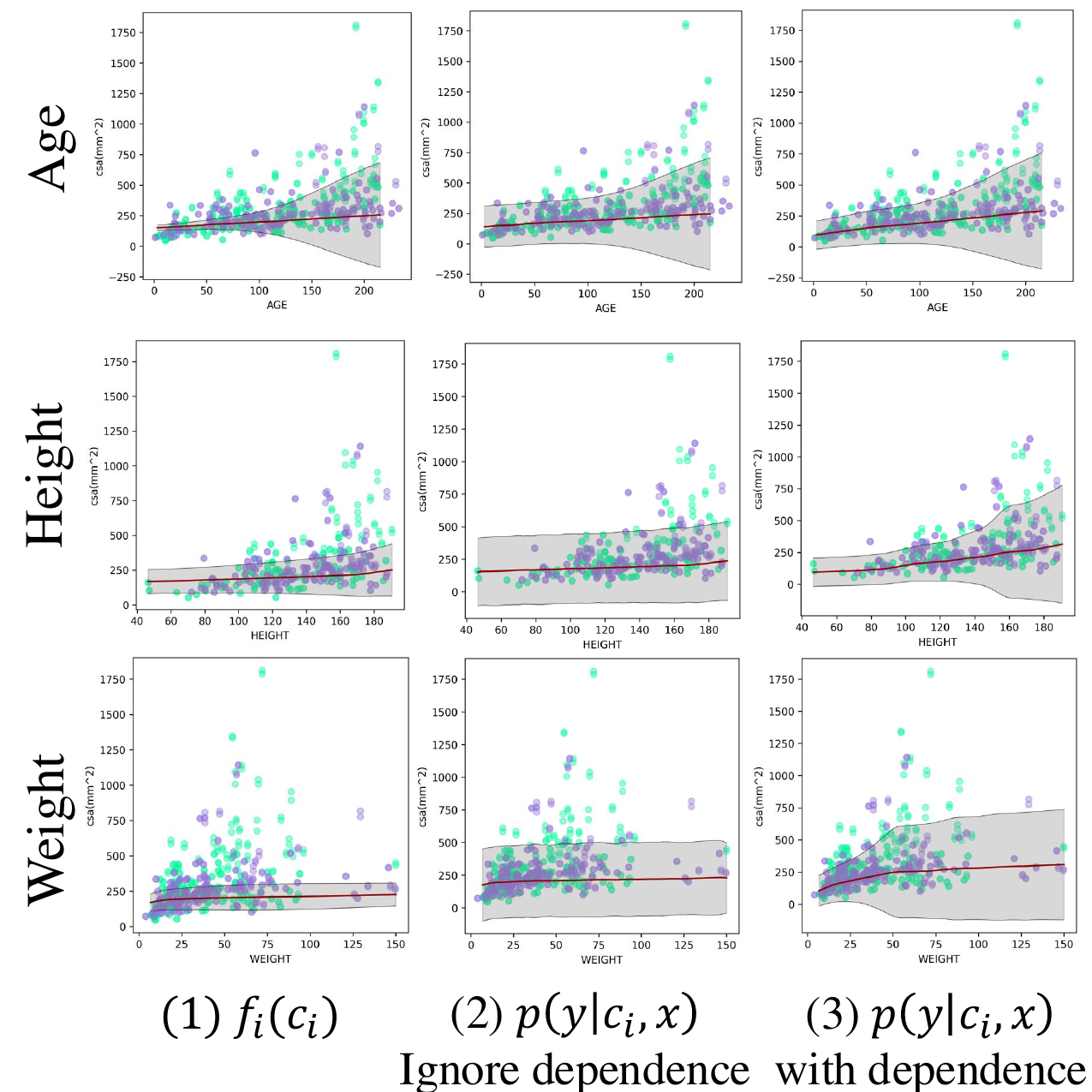}
    \caption{\small Visualizations of Covariate Interpretations from \texttt{LucidAtlas} for CSA Distribution at the Nasal Spine Landmark (Pediatric Airway Dataset). (1) $f_i(c_i)$ represents the disentangled covariate effect directly from a NAM as illustrated in Sec.~\ref{sec.dist_cov_effects}; (2) Marginalized covariate interpretation without accounting for covariate dependence; (3) Marginalized covariate interpretation incorporating covariate dependence. \textcolor{SeaGreen}{Green} dots denote training samples, while \textcolor{Orchid}{purple} dots indicate testing samples. The \textcolor{BrickRed}{red} lines represent the learned population trend, and the \textcolor{Gray}{gray} shading spans \(\pm 2 \times\) standard deviations. Considering covariate dependence is essential for accurately capturing how each covariate influences the population trend and associated uncertainties. } 
    \label{supp.fig.vis_whether_do_correlation_nasal}
\end{figure}

Fig.\ref{supp.fig.vis_whether_do_correlation_sgs}, Fig.\ref{supp.fig.vis_whether_do_correlation_epg} and Fig.\ref{supp.fig.vis_whether_do_correlation_nasal} illustrate covariate interpretations at the subglottis, epiglottic tip and nasal spine in the pediatric airway dataset. These visualizations emphasize the importance of modeling covariate dependencies when interpreting covariate effects.

\begin{figure}[t]
    \centering
    \includegraphics[width=0.97\linewidth]{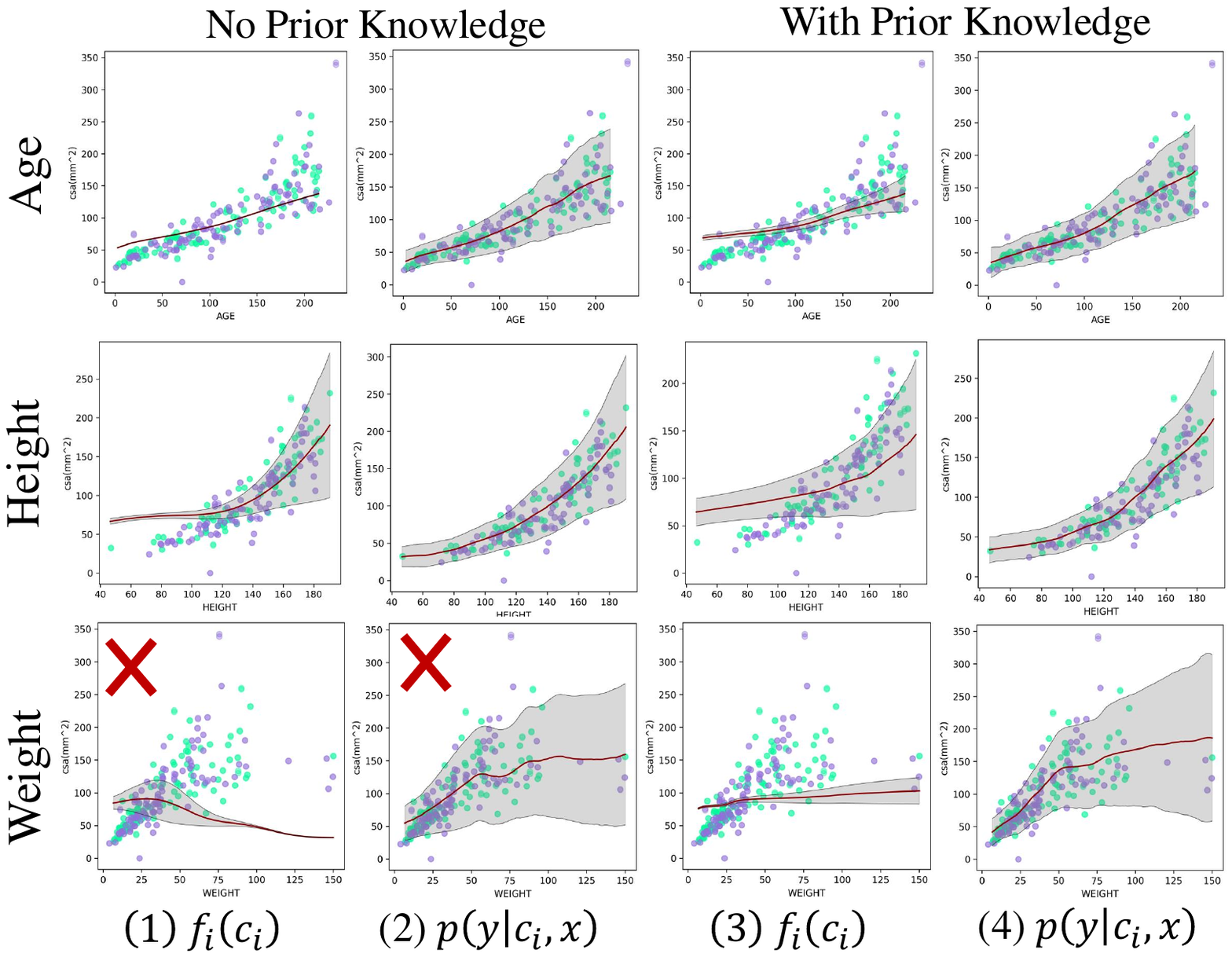}
    \caption{\small Visualizations of the Effect of Prior Knowledge in \texttt{LucidAtlas} at the Subglottis Landmark (Pediatric Airway Dataset). The \textcolor{red}{$\times$} symbol indicates the covariate interpretation contradicts prior knowledge, such as the NAM incorrectly interpreting airway CSA as decreasing with a child's weight. Without incorporating prior knowledge, the model may deviate form our prior assumptions. Without marginalization, to account for covariate dependencies, the data may not be fit well.}
    \label{supp.fig.whether_use_prior}
\end{figure}

\begin{figure}
    \centering
    \includegraphics[width=0.8\linewidth]{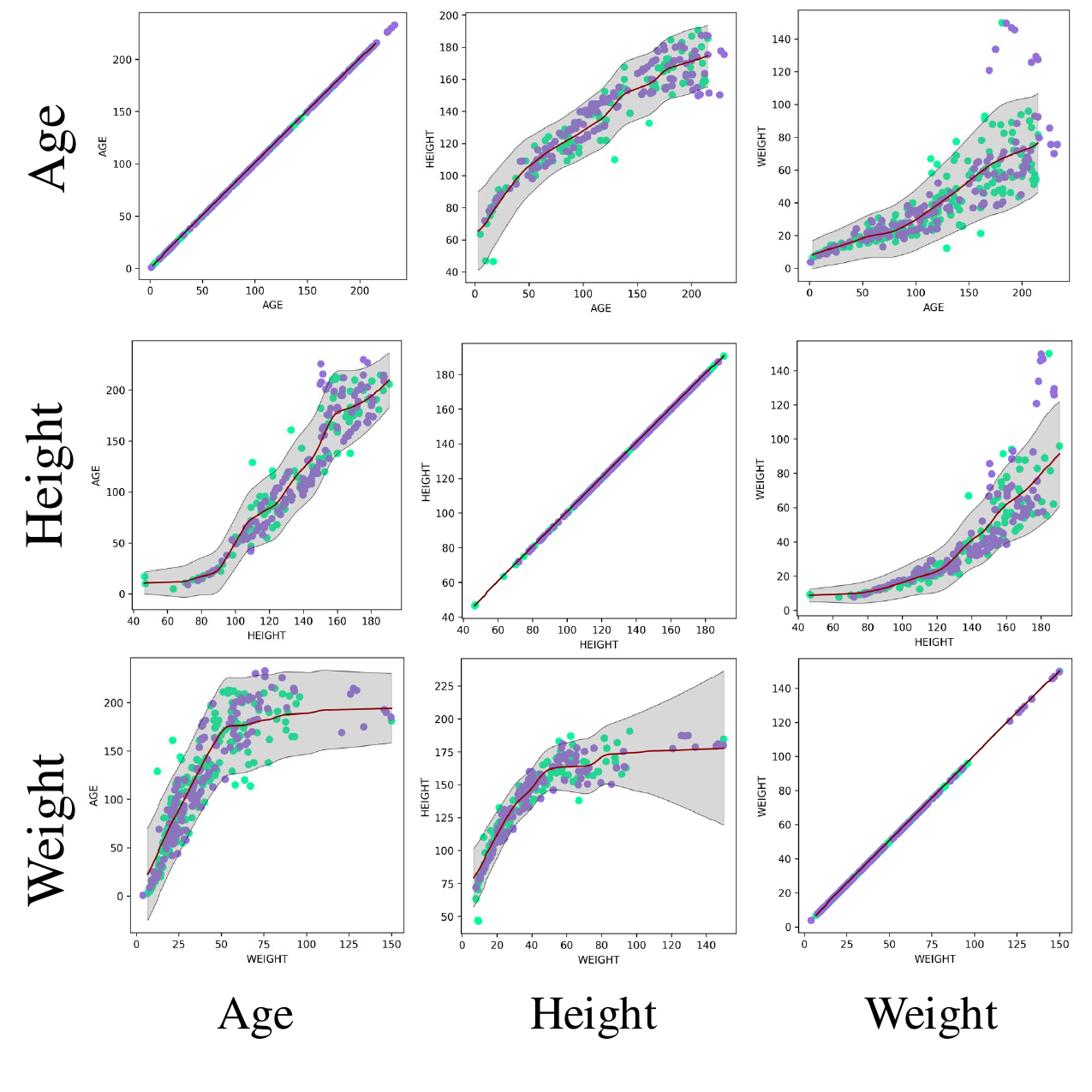}
    \caption{Pairwise Conditional Distribution $p(c_k|c_i)$ of Age, Height and Weight in the Pediatric Airway Dataset learned by \texttt{LucidAtlas}. }
    \label{supp.pairwise_cov_corr_airway}
\end{figure}

\begin{figure}
    \centering
    \includegraphics[width=0.9\linewidth]{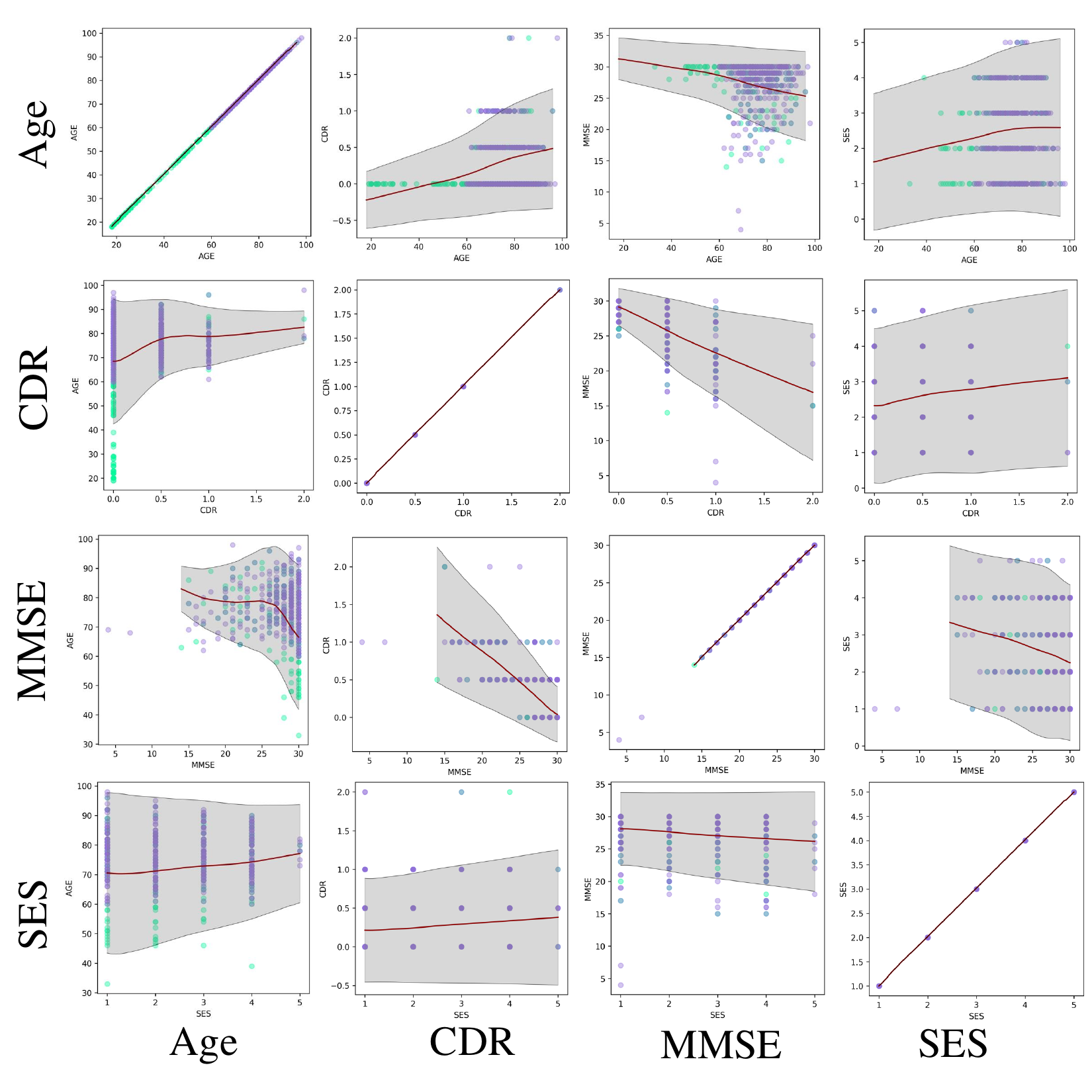}
    \caption{Pairwise Conditional Distribution $p(c_k|c_i)$ of Age, Clinical Dementia Rating (CDR) Socioeconomic Status (SES), Mini-Mental State Examination (MMSE)in the OASIS Brain Dataset learned by \texttt{LucidAtlas}. }
    \label{supp.pairwise_cov_corr_brain}
\end{figure}

Fig.~\ref{supp.fig.whether_use_prior} compares the covariate interpretations when ignoring or using covariate dependencies. 
Fig.\ref{supp.pairwise_cov_corr_airway} and Fig.\ref{supp.pairwise_cov_corr_brain} visualize the pairwise conditional distribution $p(c_k|c_i)$ for covariates in the pediatric airway dataset and the OASIS Brain dataset, respectively.

\begin{figure}
    \centering
    \includegraphics[width=1.0\linewidth]{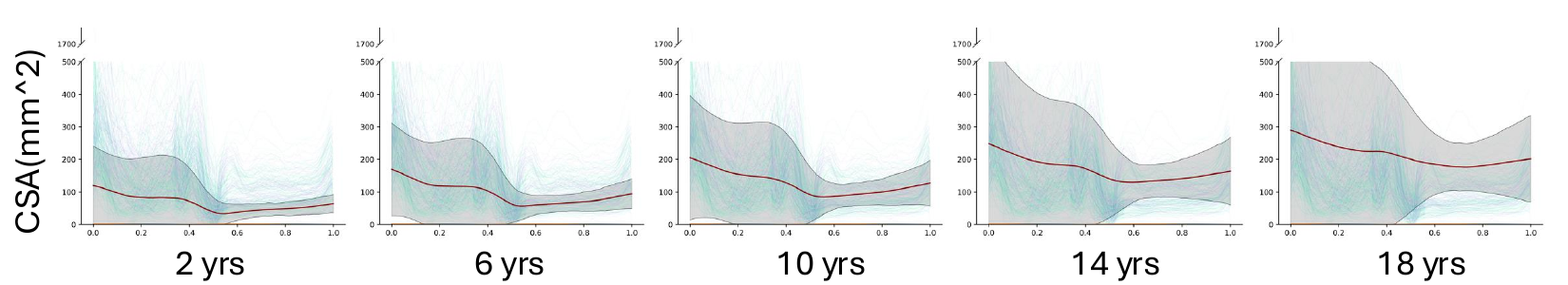}
    \caption{\small Visualization of Pediatric Airway CSA Functions with Uncertainties Across Different Ages.
In each subplot, the x-axis represents the normalized airway depth along the centerline (from the tip of the nose at 0 to the carina at 1), while the y-axis denotes the airway CSA at that depth. \textcolor{SeaGreen}{Green} and \textcolor{Orchid}{purple} lines indicate training and testing samples respectively. Both the average airway CSA and population variance increase as children grow. }
    \label{supp.fig.vis_airway_shape}
\end{figure}

Fig.~\ref{supp.fig.vis_airway_shape} visualizes pediatric airway CSA functions with uncertainties across different ages, incorporating marginalized covariate interpretation while accounting for covariate dependence in \texttt{LucidAtlas}. The visualization reveals that both the average airway CSA and population variance increase as children grow.

\begin{table*}[t]
\centering
\resizebox{0.99\textwidth}{!}{
\begin{tabular}{c|ccc|c|ccccccc}
\hline
 &
   &
   &
   &
  \cellcolor[HTML]{E9FEE9} &
  \multicolumn{7}{c}{\cellcolor[HTML]{E2E3F8}\textbf{Pediatric Airway}} \\ \cline{6-12} 
\multirow{-2}{*}{Methods} &
  \multirow{-2}{*}{Spa.} &
  \multirow{-2}{*}{Add.} &
  \multirow{-2}{*}{Mono.} &
  \multirow{-2}{*}{\cellcolor[HTML]{E9FEE9}\textbf{OASIS Brain}} &
  \multicolumn{1}{c|}{\textbf{Overall}} &
  nasal spine &
  choana &
  epiglottic tip &
  \cellcolor[HTML]{F8E5E4}TVC &
  \cellcolor[HTML]{F8E5E4}subglottis &
  \cellcolor[HTML]{F8E5E4}carina \\ \hline
MLP+NLL &
  \XSolidBrush &
  \XSolidBrush &
  \XSolidBrush &
  \textbf{0.0150} &
  \multicolumn{1}{c|}{0.0454} &
  0.0713 &
  0.052 &
  0.0727 &
  0.1023 &
  0.1071 &
  \textbf{0.0262} \\
NAMLSS &
  \XSolidBrush &
  \Checkmark &
  \XSolidBrush &
  {\color[HTML]{000000} 0.0189} &
  \multicolumn{1}{c|}{0.0572} &
  {\color[HTML]{000000} 0.116} &
  0.0316 &
  {\color[HTML]{9A0000} \textbf{0.0243}} &
  {\color[HTML]{000000} 0.1047} &
  {\color[HTML]{000000} 0.0949} &
  {\color[HTML]{000000} 0.158} \\ \hline
\rowcolor[HTML]{EFEFEF} 
Ours  np &
  \Checkmark &
  \Checkmark &
  \XSolidBrush &
  {\color[HTML]{9A0000} \textbf{0.0110}} &
  \multicolumn{1}{c|}{\cellcolor[HTML]{EFEFEF}{\color[HTML]{000000} 0.0400}} &
  {\color[HTML]{9A0000} \textbf{0.0261}} &
  {\color[HTML]{9A0000} \textbf{0.0131}} &
  0.0673 &
  {\color[HTML]{000000} \textbf{0.0686}} &
  {\color[HTML]{9A0000} \textbf{0.0452}} &
  {\color[HTML]{000000} 0.0398} \\
\rowcolor[HTML]{EFEFEF} 
Our part &
  \Checkmark &
  \Checkmark &
  \Checkmark &
  {\color[HTML]{000000} 0.0161} &
  \multicolumn{1}{c|}{\cellcolor[HTML]{EFEFEF}{\color[HTML]{9A0000} \textbf{0.0236}}} &
  \textbf{0.0266} &
  \textbf{0.0312} &
  0.1177 &
  {\color[HTML]{9A0000} \textbf{0.0545}} &
  \textbf{0.0467} &
  {\color[HTML]{9A0000} \textbf{0.0175}} \\
\rowcolor[HTML]{EFEFEF} 
Ours imp &
  \Checkmark &
  \Checkmark &
  \Checkmark &
  0.0187 &
  \multicolumn{1}{c|}{\cellcolor[HTML]{EFEFEF}{\color[HTML]{000000} \textbf{0.0248}}} &
  0.0805 &
  0.0326 &
  \textbf{0.0553} &
  0.0732 &
  0.0673 &
  {\color[HTML]{000000} 0.0522} \\ \hline
\end{tabular}
}
\caption{\small Quantitative Evaluation of Population Distribution Estimation Using Expected Calibration Error (ECE). \textbf{\textcolor{purple}{Bold red values}} indicate the best scores across all methods. \textbf{Bold black values} indicate the 2nd best scores of all methods. Our approach 
achieves the best performance overall.  }
\label{supp.exp.single_quant_ece}
\end{table*}

\begin{table}[t]
\centering
\resizebox{0.2\textwidth}{!}{
\begin{tabular}{lcl}
\hline
\rowcolor[HTML]{E9FEE9} 
\multicolumn{3}{c}{\cellcolor[HTML]{E9FEE9}\textbf{OASIS Brain}}                                                                                    \\ \hline
\multicolumn{1}{c|}{Covariate}                   & \multicolumn{1}{c|}{Corr.}                              & \multicolumn{1}{c}{Overall}            \\ \hline
\multicolumn{1}{l|}{AGE}                         & \multicolumn{1}{c|}{\XSolidBrush}                       & 1.0233                                 \\
\rowcolor[HTML]{EFEFEF} 
\multicolumn{1}{l|}{\cellcolor[HTML]{EFEFEF}AGE} & \multicolumn{1}{c|}{\cellcolor[HTML]{EFEFEF}\Checkmark} & {\color[HTML]{9A0000} \textbf{0.9151}} \\ \hline
\multicolumn{1}{l|}{SES}                         & \multicolumn{1}{c|}{\XSolidBrush}                       & 1.9071                                 \\
\rowcolor[HTML]{EFEFEF} 
\multicolumn{1}{l|}{\cellcolor[HTML]{EFEFEF}SES} & \multicolumn{1}{c|}{\cellcolor[HTML]{EFEFEF}\Checkmark} & {\color[HTML]{9A0000} \textbf{1.2447}} \\ \hline
\multicolumn{1}{l|}{MMSE}                        & \multicolumn{1}{c|}{\XSolidBrush}                       & 1.8732                                 \\
\rowcolor[HTML]{EFEFEF} 
\multicolumn{1}{l|}{\cellcolor[HTML]{EFEFEF}MMSE} & \multicolumn{1}{c|}{\cellcolor[HTML]{EFEFEF}\Checkmark} & {\color[HTML]{9A0000} \textbf{1.0849}} \\ \hline
\multicolumn{1}{l|}{CDR}                         & \multicolumn{1}{c|}{\XSolidBrush}                       & 1.8208                                 \\
\rowcolor[HTML]{EFEFEF} 
\multicolumn{1}{l|}{\cellcolor[HTML]{EFEFEF}CDR} & \multicolumn{1}{c|}{\cellcolor[HTML]{EFEFEF}\Checkmark} & {\color[HTML]{9A0000} \textbf{1.0408}} \\ \hline
\end{tabular}
}
\caption{\small Quantitative Comparison of Different Ways of Marginalization for OASIS Brain Dataset. NLL is computed between the marginalized covariate interpretation and the data distribution. A \Checkmark in the \textbf{Corr.} column indicates that covariate dependence is considered, while \XSolidBrush signifies that it is ignored. Accounting for covariate dependence improves alignment between covariate interpretation and the data distribution.
}
\label{supp.exp.oasis_cov_corr}
\end{table}

\clearpage

%% file: uai2025-template.bbl
\begin{thebibliography}{58}
\providecommand{\natexlab}[1]{#1}
\providecommand{\url}[1]{\texttt{#1}}
\expandafter\ifx\csname urlstyle\endcsname\relax
  \providecommand{\doi}[1]{doi: #1}\else
  \providecommand{\doi}{doi: \begingroup \urlstyle{rm}\Url}\fi

\bibitem[Agarwal et~al.(2020)Agarwal, Frosst, Zhang, Caruana, and Hinton]{agarwal2020neural}
Rishabh Agarwal, Nicholas Frosst, Xuezhou Zhang, Rich Caruana, and Geoffrey~E Hinton.
\newblock Neural additive models: Interpretable machine learning with neural nets.
\newblock \emph{arXiv preprint arXiv:2004.13912}, 2020.

\bibitem[Arik and Pfister(2021)]{arik2021tabnet}
Sercan~{\"O} Arik and Tomas Pfister.
\newblock Tabnet: Attentive interpretable tabular learning.
\newblock In \emph{Proceedings of the AAAI Conference on Artificial Intelligence}, volume~35, pages 6679--6687, 2021.

\bibitem[Bercea et~al.(2022)Bercea, Wiestler, Rueckert, and Albarqouni]{bercea2022federated}
Cosmin~I Bercea, Benedikt Wiestler, Daniel Rueckert, and Shadi Albarqouni.
\newblock Federated disentangled representation learning for unsupervised brain anomaly detection.
\newblock \emph{Nature Machine Intelligence}, 4\penalty0 (8):\penalty0 685--695, 2022.

\bibitem[Berrevoets et~al.(2021)Berrevoets, Alaa, Qian, Jordon, Gimson, and Van Der~Schaar]{berrevoets2021learning}
Jeroen Berrevoets, Ahmed Alaa, Zhaozhi Qian, James Jordon, Alexander~ES Gimson, and Mihaela Van Der~Schaar.
\newblock Learning queueing policies for organ transplantation allocation using interpretable counterfactual survival analysis.
\newblock In \emph{International Conference on Machine Learning}, pages 792--802. PMLR, 2021.

\bibitem[Bouchiat et~al.(2023)Bouchiat, Immer, Y{\`e}che, R{\"a}tsch, and Fortuin]{bouchiat2023lanam}
Kouroche Bouchiat, Alexander Immer, Hugo Y{\`e}che, Gunnar R{\"a}tsch, and Vincent Fortuin.
\newblock Improving neural additive models with bayesian principles.
\newblock \emph{arXiv preprint arXiv:2305.16905}, 2023.

\bibitem[Chartsias et~al.(2019)Chartsias, Joyce, Papanastasiou, Semple, Williams, Newby, Dharmakumar, and Tsaftaris]{chartsias2019disentangled}
Agisilaos Chartsias, Thomas Joyce, Giorgos Papanastasiou, Scott Semple, Michelle Williams, David~E Newby, Rohan Dharmakumar, and Sotirios~A Tsaftaris.
\newblock Disentangled representation learning in cardiac image analysis.
\newblock \emph{Medical image analysis}, 58:\penalty0 101535, 2019.

\bibitem[Chen et~al.(2022)Chen, Yin, and Long]{chen2022covariate}
Kan Chen, Qishuo Yin, and Qi~Long.
\newblock Covariate-balancing-aware interpretable deep learning models for treatment effect estimation.
\newblock \emph{arXiv preprint arXiv:2203.03185}, 2022.

\bibitem[Chu et~al.(2022)Chu, Zhang, Huang, Si, Huang, and Huang]{chu2022disentangled}
Jiebin Chu, Yaoyun Zhang, Fei Huang, Luo Si, Songfang Huang, and Zhengxing Huang.
\newblock Disentangled representation for sequential treatment effect estimation.
\newblock \emph{Computer Methods and Programs in Biomedicine}, 226:\penalty0 107175, 2022.

\bibitem[Commowick et~al.(2005)Commowick, Stefanescu, Fillard, Arsigny, Ayache, Pennec, and Malandain]{commowick2005incorporating}
Olivier Commowick, Radu Stefanescu, Pierre Fillard, Vincent Arsigny, Nicholas Ayache, Xavier Pennec, and Gr{\'e}goire Malandain.
\newblock Incorporating statistical measures of anatomical variability in atlas-to-subject registration for conformal brain radiotherapy.
\newblock In \emph{International Conference on Medical Image Computing and Computer-Assisted Intervention}, pages 927--934. Springer, 2005.

\bibitem[Crabbe et~al.(2021)Crabbe, Qian, Imrie, and van~der Schaar]{Crabbe2021Simplex}
Jonathan Crabbe, Zhaozhi Qian, Fergus Imrie, and Mihaela van~der Schaar.
\newblock Explaining latent representations with a corpus of examples.
\newblock In M.~Ranzato, A.~Beygelzimer, Y.~Dauphin, P.S. Liang, and J.~Wortman Vaughan, editors, \emph{Advances in Neural Information Processing Systems}, volume~34, pages 12154--12166. Curran Associates, Inc., 2021.
\newblock URL \url{https://proceedings.neurips.cc/paper/2021/file/65658fde58ab3c2b6e5132a39fae7cb9-Paper.pdf}.

\bibitem[Daxberger et~al.(2021)Daxberger, Kristiadi, Immer, Eschenhagen, Bauer, and Hennig]{daxberger2021laplace}
Erik Daxberger, Agustinus Kristiadi, Alexander Immer, Runa Eschenhagen, Matthias Bauer, and Philipp Hennig.
\newblock Laplace redux-effortless bayesian deep learning.
\newblock \emph{Advances in Neural Information Processing Systems}, 34:\penalty0 20089--20103, 2021.

\bibitem[De~Cao et~al.(2020)De~Cao, Aziz, and Titov]{de2020block}
Nicola De~Cao, Wilker Aziz, and Ivan Titov.
\newblock Block neural autoregressive flow.
\newblock In \emph{Uncertainty in artificial intelligence}, pages 1263--1273. PMLR, 2020.

\bibitem[Ding et~al.(2020)Ding, Xu, Xu, Parmar, Yang, Welling, and Tu]{ding2020guided}
Zheng Ding, Yifan Xu, Weijian Xu, Gaurav Parmar, Yang Yang, Max Welling, and Zhuowen Tu.
\newblock Guided variational autoencoder for disentanglement learning.
\newblock In \emph{Proceedings of the IEEE/CVF conference on computer vision and pattern recognition}, pages 7920--7929, 2020.

\bibitem[Fotenos et~al.(2008)Fotenos, Mintun, Snyder, Morris, and Buckner]{fotenos2008brain}
Anthony~F Fotenos, Mark~A Mintun, Abraham~Z Snyder, John~C Morris, and Randy~L Buckner.
\newblock Brain volume decline in aging: evidence for a relation between socioeconomic status, preclinical {Alzheimer} disease, and reserve.
\newblock \emph{Archives of neurology}, 65\penalty0 (1):\penalty0 113--120, 2008.

\bibitem[Friedman(2001)]{friedman2001greedy}
Jerome~H Friedman.
\newblock Greedy function approximation: a gradient boosting machine.
\newblock \emph{Annals of statistics}, pages 1189--1232, 2001.

\bibitem[Gal and Ghahramani(2016)]{gal2016dropout}
Yarin Gal and Zoubin Ghahramani.
\newblock Dropout as a bayesian approximation: Representing model uncertainty in deep learning.
\newblock In \emph{international conference on machine learning}, pages 1050--1059. PMLR, 2016.

\bibitem[Hastie(2017)]{hastie2017generalized}
Trevor~J Hastie.
\newblock \emph{Generalized additive models}.
\newblock Routledge, 2017.

\bibitem[Hong et~al.(2013)Hong, Niethammer, Andruejol, Kimbell, Pitkin, Superfine, Davis, Zdanski, and Davis]{atlashong2013pediatric}
Yi~Hong, Marc Niethammer, Johan Andruejol, Julia~S Kimbell, Elizabeth Pitkin, Richard Superfine, Stephanie Davis, Carlton~J Zdanski, and Brad Davis.
\newblock A pediatric airway atlas and its application in subglottic stenosis.
\newblock In \emph{2013 Ieee 10th International Symposium on Biomedical Imaging}, pages 1206--1209. IEEE, 2013.

\bibitem[H{\"u}llermeier and Waegeman(2021)]{hullermeier2021survey}
Eyke H{\"u}llermeier and Willem Waegeman.
\newblock Aleatoric and epistemic uncertainty in machine learning: An introduction to concepts and methods.
\newblock \emph{Machine learning}, 110\penalty0 (3):\penalty0 457--506, 2021.

\bibitem[Immer et~al.(2024)Immer, Palumbo, Marx, and Vogt]{immer2024effectiveheterbayes}
Alexander Immer, Emanuele Palumbo, Alexander Marx, and Julia Vogt.
\newblock Effective bayesian heteroscedastic regression with deep neural networks.
\newblock \emph{Advances in Neural Information Processing Systems}, 36, 2024.

\bibitem[Jack~Jr et~al.(2008)Jack~Jr, Bernstein, Fox, Thompson, Alexander, Harvey, Borowski, Britson, L.~Whitwell, Ward, et~al.]{jack2008alzheimer}
Clifford~R Jack~Jr, Matt~A Bernstein, Nick~C Fox, Paul Thompson, Gene Alexander, Danielle Harvey, Bret Borowski, Paula~J Britson, Jennifer L.~Whitwell, Chadwick Ward, et~al.
\newblock The {Alzheimer's} disease neuroimaging initiative ({ADNI}): {MRI} methods.
\newblock \emph{Journal of Magnetic Resonance Imaging: An Official Journal of the International Society for Magnetic Resonance in Medicine}, 27\penalty0 (4):\penalty0 685--691, 2008.

\bibitem[Jiao et~al.(2023)Jiao, Zdanski, Kimbell, Prince, Worden, Kirse, Rutter, Shields, Dunn, Mahmud, et~al.]{jiao2023naisr}
Yining Jiao, Carlton Zdanski, Julia Kimbell, Andrew Prince, Cameron Worden, Samuel Kirse, Christopher Rutter, Benjamin Shields, William Dunn, Jisan Mahmud, et~al.
\newblock Naisr: A {3D} neural additive model for interpretable shape representation.
\newblock \emph{arXiv preprint arXiv:2303.09234}, 2023.

\bibitem[Jin et~al.(2019)Jin, Udupa, and Torigian]{jin2019howmany}
Ze~Jin, Jayaram~K Udupa, and Drew~A Torigian.
\newblock How many models/atlases are needed as priors for capturing anatomic population variations?
\newblock \emph{Medical image analysis}, 58:\penalty0 101550, 2019.

\bibitem[John et~al.(2018)John, Mou, Bahuleyan, and Vechtomova]{john2018disentangled}
Vineet John, Lili Mou, Hareesh Bahuleyan, and Olga Vechtomova.
\newblock Disentangled representation learning for non-parallel text style transfer.
\newblock \emph{arXiv preprint arXiv:1808.04339}, 2018.

\bibitem[Joshi et~al.(2004)Joshi, Davis, Jomier, and Gerig]{joshi2004unbiased}
Sarang Joshi, Brad Davis, Matthieu Jomier, and Guido Gerig.
\newblock Unbiased diffeomorphic atlas construction for computational anatomy.
\newblock \emph{NeuroImage}, 23:\penalty0 S151--S160, 2004.

\bibitem[Jung and Oh(2021)]{jung2021towards}
Hyungsik Jung and Youngrock Oh.
\newblock Towards better explanations of class activation mapping.
\newblock In \emph{Proceedings of the IEEE/CVF International Conference on Computer Vision}, pages 1336--1344, 2021.

\bibitem[Ke et~al.(2017)Ke, Meng, Finley, Wang, Chen, Ma, Ye, and Liu]{ke2017lightgbm}
Guolin Ke, Qi~Meng, Thomas Finley, Taifeng Wang, Wei Chen, Weidong Ma, Qiwei Ye, and Tie-Yan Liu.
\newblock Lightgbm: A highly efficient gradient boosting decision tree.
\newblock \emph{Advances in neural information processing systems}, 30, 2017.

\bibitem[Kitouni et~al.(2023)Kitouni, Nolte, and Williams]{kitouni2023mono}
Ouail Kitouni, Niklas Nolte, and Michael Williams.
\newblock Expressive monotonic neural networks.
\newblock \emph{arXiv preprint arXiv:2307.07512}, 2023.

\bibitem[Kova{\v{c}}evi{\'c} et~al.(2005)Kova{\v{c}}evi{\'c}, Henderson, Chan, Lifshitz, Bishop, Evans, Henkelman, and Chen]{kovavcevic2005three}
N~Kova{\v{c}}evi{\'c}, JT~Henderson, E~Chan, N~Lifshitz, J~Bishop, AC~Evans, RM~Henkelman, and XJ~Chen.
\newblock A three-dimensional {MRI} atlas of the mouse brain with estimates of the average and variability.
\newblock \emph{Cerebral cortex}, 15\penalty0 (5):\penalty0 639--645, 2005.

\bibitem[Lakshminarayanan et~al.(2017)Lakshminarayanan, Pritzel, and Blundell]{lakshminarayanan2017deepemsemble}
Balaji Lakshminarayanan, Alexander Pritzel, and Charles Blundell.
\newblock Simple and scalable predictive uncertainty estimation using deep ensembles.
\newblock \emph{Advances in neural information processing systems}, 30, 2017.

\bibitem[Liu et~al.(2020)Liu, Han, Zhang, and Liu]{liu2020certified}
Xingchao Liu, Xing Han, Na~Zhang, and Qiang Liu.
\newblock Certified monotonic neural networks.
\newblock \emph{Advances in Neural Information Processing Systems}, 33:\penalty0 15427--15438, 2020.

\bibitem[Lou et~al.(2013)Lou, Caruana, Gehrke, and Hooker]{lou2013EBM}
Yin Lou, Rich Caruana, Johannes Gehrke, and Giles Hooker.
\newblock Accurate intelligible models with pairwise interactions.
\newblock In \emph{Proceedings of the 19th ACM SIGKDD international conference on Knowledge discovery and data mining}, pages 623--631, 2013.

\bibitem[Lu et~al.(2022)Lu, Boukouvalas, and Hensman]{lu2022oak}
Xiaoyu Lu, Alexis Boukouvalas, and James Hensman.
\newblock Additive gaussian processes revisited.
\newblock In \emph{International Conference on Machine Learning}, pages 14358--14383. PMLR, 2022.

\bibitem[Lundberg(2017)]{lundberg2017unified}
Scott Lundberg.
\newblock A unified approach to interpreting model predictions.
\newblock \emph{arXiv preprint arXiv:1705.07874}, 2017.

\bibitem[Marcus et~al.(2007)Marcus, Wang, Parker, Csernansky, Morris, and Buckner]{marcus2007oasis}
Daniel~S Marcus, Tracy~H Wang, Jamie Parker, John~G Csernansky, John~C Morris, and Randy~L Buckner.
\newblock Open access series of imaging studies (oasis): cross-sectional {MRI} data in young, middle aged, nondemented, and demented older adults.
\newblock \emph{Journal of cognitive neuroscience}, 19\penalty0 (9):\penalty0 1498--1507, 2007.

\bibitem[Moraffah et~al.(2020)Moraffah, Karami, Guo, Raglin, and Liu]{moraffah2020causal}
Raha Moraffah, Mansooreh Karami, Ruocheng Guo, Adrienne Raglin, and Huan Liu.
\newblock Causal interpretability for machine learning-problems, methods and evaluation.
\newblock \emph{ACM SIGKDD Explorations Newsletter}, 22\penalty0 (1):\penalty0 18--33, 2020.

\bibitem[Nori et~al.(2019)Nori, Jenkins, Koch, and Caruana]{nori2019interpretmlunifiedframeworkmachine}
Harsha Nori, Samuel Jenkins, Paul Koch, and Rich Caruana.
\newblock Interpretml: A unified framework for machine learning interpretability, 2019.
\newblock URL \url{https://arxiv.org/abs/1909.09223}.

\bibitem[Ribeiro et~al.(2016)Ribeiro, Singh, and Guestrin]{ribeiro2016should}
Marco~Tulio Ribeiro, Sameer Singh, and Carlos Guestrin.
\newblock "{W}hy should i trust you?" {Explaining} the predictions of any classifier.
\newblock In \emph{Proceedings of the 22nd ACM SIGKDD international conference on knowledge discovery and data mining}, pages 1135--1144, 2016.

\bibitem[Ronneberger et~al.(2015)Ronneberger, Fischer, and Brox]{unetronneberger2015u}
Olaf Ronneberger, Philipp Fischer, and Thomas Brox.
\newblock U-net: Convolutional networks for biomedical image segmentation.
\newblock In \emph{Medical image computing and computer-assisted intervention--MICCAI 2015: 18th international conference, Munich, Germany, October 5-9, 2015, proceedings, part III 18}, pages 234--241. Springer, 2015.

\bibitem[Rudin(2019)]{rudin2019stop}
Cynthia Rudin.
\newblock Stop explaining black box machine learning models for high stakes decisions and use interpretable models instead.
\newblock \emph{Nature machine intelligence}, 1\penalty0 (5):\penalty0 206--215, 2019.

\bibitem[Runje and Shankaranarayana(2023)]{runje2023constrained}
Davor Runje and Sharath~M Shankaranarayana.
\newblock Constrained monotonic neural networks.
\newblock In \emph{International Conference on Machine Learning}, pages 29338--29353. PMLR, 2023.

\bibitem[Shoshan et~al.(2021)Shoshan, Bhonker, Kviatkovsky, and Medioni]{shoshan2021gan}
Alon Shoshan, Nadav Bhonker, Igor Kviatkovsky, and Gerard Medioni.
\newblock {GAN}-control: Explicitly controllable {GAN}s.
\newblock In \emph{Proceedings of the IEEE/CVF international conference on computer vision}, pages 14083--14093, 2021.

\bibitem[Siems et~al.(2023)Siems, Ditschuneit, Ripken, Lindborg, Schambach, Otterbach, and Genzel]{siems2023curve}
Julien Siems, Konstantin Ditschuneit, Winfried Ripken, Alma Lindborg, Maximilian Schambach, Johannes Otterbach, and Martin Genzel.
\newblock Curve your enthusiasm: concurvity regularization in differentiable generalized additive models.
\newblock \emph{Advances in Neural Information Processing Systems}, 36:\penalty0 19029--19057, 2023.

\bibitem[Stirn et~al.(2022)Stirn, Wessels, Schertzer, Pereira, Sanjana, and Knowles]{stirn2022faithfulheteroscedasticregressionneural}
Andrew Stirn, Hans-Hermann Wessels, Megan Schertzer, Laura Pereira, Neville~E. Sanjana, and David~A. Knowles.
\newblock Faithful heteroscedastic regression with neural networks, 2022.
\newblock URL \url{https://arxiv.org/abs/2212.09184}.

\bibitem[Thiagarajan et~al.(2020)Thiagarajan, Sattigeri, Rajan, and Venkatesh]{thiagarajan2020calibrating}
Jayaraman~J Thiagarajan, Prasanna Sattigeri, Deepta Rajan, and Bindya Venkatesh.
\newblock Calibrating healthcare {AI}: Towards reliable and interpretable deep predictive models.
\newblock \emph{arXiv preprint arXiv:2004.14480}, 2020.

\bibitem[Thielmann et~al.(2024)Thielmann, Kruse, Kneib, and S{\"a}fken]{thielmann2024namlss}
Anton~Frederik Thielmann, Ren{\'e}-Marcel Kruse, Thomas Kneib, and Benjamin S{\"a}fken.
\newblock Neural additive models for location scale and shape: A framework for interpretable neural regression beyond the mean.
\newblock In \emph{International Conference on Artificial Intelligence and Statistics}, pages 1783--1791. PMLR, 2024.

\bibitem[Thompson and Toga(2002)]{thompson2002framework}
Paul~M Thompson and Arthur~W Toga.
\newblock A framework for computational anatomy.
\newblock \emph{Computing and Visualization in Science}, 5\penalty0 (1):\penalty0 13--34, 2002.

\bibitem[Wang et~al.(2025)Wang, Wang, Zhou, Peng, Song, Zhang, Sun, Niu, Liu, Chen, et~al.]{wang2025uncertaintysurvey}
Tianyang Wang, Yunze Wang, Jun Zhou, Benji Peng, Xinyuan Song, Charles Zhang, Xintian Sun, Qian Niu, Junyu Liu, Silin Chen, et~al.
\newblock From aleatoric to epistemic: Exploring uncertainty quantification techniques in artificial intelligence.
\newblock \emph{arXiv preprint arXiv:2501.03282}, 2025.

\bibitem[Woo et~al.(2018)Woo, Park, Lee, and Kweon]{woo2018cbam}
Sanghyun Woo, Jongchan Park, Joon-Young Lee, and In~So Kweon.
\newblock Cbam: Convolutional block attention module.
\newblock In \emph{Proceedings of the European conference on computer vision (ECCV)}, pages 3--19, 2018.

\bibitem[Wooldridge et~al.(2016)Wooldridge, Wadud, and Lye]{wooldridge2016homohetr}
Jeffrey~M Wooldridge, Mokhtarul Wadud, and Jenny Lye.
\newblock \emph{Introductory econometrics: Asia pacific edition with online study tools 12 months}.
\newblock Cengage AU, 2016.

\bibitem[Xu et~al.(2021)Xu, Zhang, Zhou, Xu, Qi, and Qiao]{xu2021learning}
Mutian Xu, Junhao Zhang, Zhipeng Zhou, Mingye Xu, Xiaojuan Qi, and Yu~Qiao.
\newblock Learning geometry-disentangled representation for complementary understanding of 3d object point cloud.
\newblock In \emph{Proceedings of the AAAI Conference on Artificial Intelligence}, volume~35, pages 3056--3064, 2021.

\bibitem[Yang et~al.(2020)Yang, Mo, Lai, Guibas, and Gao]{yang2020dsm}
Jie Yang, Kaichun Mo, Yu-Kun Lai, Leonidas~J Guibas, and Lin Gao.
\newblock Dsm-net: Disentangled structured mesh net for controllable generation of fine geometry.
\newblock \emph{arXiv preprint arXiv:2008.05440}, 2\penalty0 (3), 2020.

\bibitem[You et~al.(2017)You, Ding, Canini, Pfeifer, and Gupta]{you2017deep}
Seungil You, David Ding, Kevin Canini, Jan Pfeifer, and Maya Gupta.
\newblock Deep lattice networks and partial monotonic functions.
\newblock \emph{Advances in neural information processing systems}, 30, 2017.

\bibitem[Zhang et~al.(2018{\natexlab{a}})Zhang, Zhang, Zhang, Tenenbaum, Freeman, and Wu]{zhang2018learning}
Xiuming Zhang, Zhoutong Zhang, Chengkai Zhang, Josh Tenenbaum, Bill Freeman, and Jiajun Wu.
\newblock Learning to reconstruct shapes from unseen classes.
\newblock \emph{Advances in neural information processing systems}, 31, 2018{\natexlab{a}}.

\bibitem[Zhang et~al.(2018{\natexlab{b}})Zhang, Guo, Jin, Luo, He, and Lee]{zhang2018unsupervised}
Yuting Zhang, Yijie Guo, Yixin Jin, Yijun Luo, Zhiyuan He, and Honglak Lee.
\newblock Unsupervised discovery of object landmarks as structural representations.
\newblock In \emph{Proceedings of the IEEE Conference on Computer Vision and Pattern Recognition}, pages 2694--2703, 2018{\natexlab{b}}.

\bibitem[Zhou et~al.(2016)Zhou, Khosla, Lapedriza, Oliva, and Torralba]{zhou2016cvpr}
Bolei Zhou, Aditya Khosla, Agata Lapedriza, Aude Oliva, and Antonio Torralba.
\newblock Learning deep features for discriminative localization.
\newblock In \emph{Computer Vision and Pattern Recognition}, 2016.

\bibitem[Zhou and Wei(2020)]{zhou2020identifiability}
Ding Zhou and Xue-Xin Wei.
\newblock Learning identifiable and interpretable latent models of high-dimensional neural activity using pi-vae.
\newblock \emph{Advances in Neural Information Processing Systems}, 33:\penalty0 7234--7247, 2020.

\bibitem[Özgün Çiçek et~al.(2016)Özgün Çiçek, Abdulkadir, Lienkamp, Brox, and Ronneberger]{unetcciccek20163d}
Özgün Çiçek, Ahmed Abdulkadir, Soeren~S. Lienkamp, Thomas Brox, and Olaf Ronneberger.
\newblock {3D} u-net: Learning dense volumetric segmentation from sparse annotation, 2016.
\newblock URL \url{https://arxiv.org/abs/1606.06650}.

\end{thebibliography}
